\documentclass[10pt,twocolumn,letterpaper]{article}
\usepackage[pagenumbers]{cvpr} 
\usepackage[dvipsnames]{xcolor}
\usepackage{times}
\usepackage{graphicx}
\usepackage{amsmath,amssymb} 
\usepackage{url}
\usepackage{cite}
\usepackage{color}
\usepackage{epsfig}
\usepackage{graphicx}
\usepackage{wrapfig}
\usepackage{amsmath}
\usepackage{amssymb}
\usepackage{multirow}
\usepackage{booktabs}
\usepackage{algorithmic}
\usepackage{arydshln}
\usepackage{threeparttable}
\usepackage{colortbl}
\usepackage{enumitem}
\usepackage{siunitx}
\usepackage{bm}
\usepackage[numbers,sort&compress]{natbib}
\usepackage{array}
\usepackage{colortbl}
\usepackage{dblfloatfix}
\usepackage{marvosym}
\usepackage[nopar]{lipsum}
\usepackage{listings}
\usepackage[export]{adjustbox}
\usepackage{caption}
\usepackage{amsfonts}
\usepackage{bbm}
\usepackage{subfig}
\usepackage{pifont}

\usepackage{mathtools}
\usepackage{siunitx}

\makeatletter
\@namedef{ver@everyshi.sty}{}
\makeatother
\usepackage{tikz}
\usetikzlibrary{backgrounds}
\usetikzlibrary{arrows,shapes}
\usetikzlibrary{tikzmark}
\usetikzlibrary{calc}

\usepackage{algorithm}
\usepackage{listings}

\usepackage{etoolbox}
\makeatletter
\AfterEndEnvironment{algorithm}{\let\@algcomment\relax}
\AtEndEnvironment{algorithm}{\kern2pt\hrule\relax\vskip3pt\@algcomment}
\let\@algcomment\relax
\newcommand\algcomment[1]{\def\@algcomment{\footnotesize#1}}
\renewcommand\fs@ruled{\def\@fs@cfont{\bfseries}\let\@fs@capt\floatc@ruled
  \def\@fs@pre{\hrule height.8pt depth0pt \kern2pt}%
  \def\@fs@post{}%
  \def\@fs@mid{\kern2pt\hrule\kern2pt}%
  \let\@fs@iftopcapt\iftrue}
\makeatother

\usepackage{mathtools, nccmath}
\usepackage{comment}
\usepackage{xspace}
\usepackage{ragged2e}
\newcolumntype{P}[1]{>{\RaggedRight\hspace{0pt}}p{#1}}
\newcolumntype{X}[1]{>{\RaggedRight\hspace*{0pt}}p{#1}}
\usepackage{tcolorbox}
\usepackage[edges]{forest}
\colorlet{linecol}{black!75}
\usepackage{xkcdcolors} 
\newcommand{\highlight}[2]{\colorbox{#1!70}{$\strut #2$}}

\colorlet{mhpurple}{Plum!80}
\renewcommand{\highlight}[2]{\colorbox{#1!70}{\strut #2}}

\newcommand{\myhyperlink}[3][black]{\hyperlink{#2}{\color{#1}{#3}}}

\usepackage[pagebackref=true,breaklinks=true,letterpaper=true,colorlinks,bookmarks=false]{hyperref}

\makeatletter
\DeclareRobustCommand\onedot{\futurelet\@let@token\@onedot}
\def\@onedot{\ifx\@let@token.\else.\null\fi\xspace}

\definecolor{mygray2}{gray}{.6}
\definecolor{mygray3}{gray}{.3}
\definecolor{mygray}{gray}{.9}
\definecolor{mywarning}{RGB}{233,144,61}
\definecolor{ggray}{RGB}{127,127,127}
\definecolor{reda}{RGB}{192,0,0}
\definecolor{redb}{RGB}{217,148,143}
\definecolor{myyellow}{RGB}{190,144,0}
\definecolor{mygreen}{RGB}{0,153,0}
\definecolor{mygreen2}{RGB}{153,255,153}
\definecolor{myred}{RGB}{177,35,15}
\definecolor{myblue}{RGB}{58,79,116}
\definecolor{myy}{RGB}{254,203,50}
\definecolor{myy2}{RGB}{191,144,0}
\definecolor{myg}{RGB}{205,205,205}
\definecolor{myg2}{RGB}{80,80,80}
\definecolor{mypurple}{RGB}{103,72,186}
\definecolor{mycodered}{RGB}{195,63,56}
\definecolor{mycodeblue}{RGB}{30,93,190}
\definecolor{mycodegreen}{RGB}{64,145,92}
\newcommand{\pub}[1]{\color{mygray3}{\tiny{[{#1}]}}}
\usepackage{tabulary}
\newcolumntype{y}[1]{>{\raggedright\arraybackslash}p{#1pt}}
\newcolumntype{z}[1]{>{\raggedleft\arraybackslash}p{#1pt}}

\makeatother
\makeatletter
\newcommand{\thickhline}{%
   \noalign {\ifnum 0=`}\fi \hrule height 1pt
   \futurelet \reserved@a \@xhline
}
\makeatother
\def\cvprPaperID{2668} 
\def\confName{CVPR}
\def\confYear{2023}

\begin{document}

\title{Unified Mask Embedding and Correspondence Learning for\\Self-Supervised Video Segmentation}

\author{Liulei Li$^{1,4}$\thanks{Work done during an internship at Baidu VIS.}~~, Wenguan Wang$^{1}$\thanks{Corresponding author: \textit{Wenguan Wang}.}~, Tianfei Zhou$^{2}$, Jianwu Li$^{3}$, Yi Yang$^{1}$\\
\small{$^1$ ReLER, CCAI, Zhejiang University}~~\small{$^2$ ETH Zurich}~~\small{$^3$ Beijing Institute of Technology}~~\small{$^4$ Baidu VIS}\\
\small\url{https://github.com/0liliulei/Mask-VOS}
}

\maketitle

\begin{abstract}
The objective of this paper is self-supervised learning~of video object segmentation.  We develop a unified framework which simultaneously models cross-frame dense correspon- dence$_{\!}$ for$_{\!}$ locally$_{\!}$ discriminative$_{\!}$ feature$_{\!}$ learning$_{\!}$ and$_{\!}$ embeds$_{\!}$ object-level context for target-mask decoding. As a result,~it is able to directly learn to perform mask-guided~sequential segmentation$_{\!}$ from$_{\!}$ \text{unlabeled}$_{\!}$ videos,$_{\!}$ in$_{\!}$ contrast$_{\!}$ to$_{\!}$ previous$_{\!}$ efforts$_{\!}$ usually$_{\!}$ relying$_{\!}$ on$_{\!}$~an$_{\!}$ \textit{oblique}$_{\!}$ solution$_{\!}$ ---$_{\!}$ cheaply ``copying''$_{\!}$ labels$_{\!}$ according$_{\!}$ to$_{\!}$ pixel-wise$_{\!}$ correlations. Concretely, our algorithm alternates between \textbf{i)} clustering video pixels for creating pseudo$_{\!}$ segmentation$_{\!}$ labels$_{\!}$ \textit{ex}$_{\!}$ \textit{nihilo};$_{\!}$ and$_{\!}$ \textbf{ii)}$_{\!}$ utilizing$_{\!}$ the$_{\!}$ pseudo$_{\!}$ labels$_{\!}$~to learn mask encoding and decoding for VOS. Unsupervised correspondence learning~is further incorporated into this self-taught, mask embedding scheme,$_{\!}$ so$_{\!}$ as$_{\!}$ to$_{\!}$ ensure$_{\!}$ the$_{\!}$ generic$_{\!}$ nature$_{\!}$ of$_{\!}$ the$_{\!}$ learnt$_{\!}$ repre- sentation and avoid cluster degeneracy. Our algorithm sets state-of-the-arts on two standard benchmarks (\ie, DAVIS$_{17}$ and YouTube-VOS), narrowing the gap between self- and fully-supervised VOS, in terms of both performance and network architecture design.
\end{abstract}

\vspace{-6pt}
\section{Introduction}
\vspace{-1pt}
$_{\!}$In this article, we focus on a classic computer vision~task: accurately segmenting desired$_{\!}$ object(s)$_{\!}$ in$_{\!}$ a$_{\!}$ video$_{\!}$ sequence, where$_{\!}$ the$_{\!}$ target$_{\!}$ object(s)$_{\!}$ are$_{\!}$ defined$_{\!}$ by$_{\!}$ pixel-wise$_{\!}$ mask(s) in$_{\!}$ the$_{\!}$ first$_{\!}$ frame.$_{\!}$ This$_{\!}$ task$_{\!}$ is$_{\!}$ referred$_{\!}$ as$_{\!}$ (\textit{one-shot})$_{\!}$ \textit{video}$_{\!}$ \textit{ob-}
\textit{ject segmentation}$_{\!}$ (VOS)$_{\!}$ or$_{\!}$ \textit{mask$_{\!}$ propagation}$_{\!}$~\cite{wang2021survey},$_{\!}$ playing$_{\!}$~a$_{\!}$
vital$_{\!}$ role$_{\!}$ in$_{\!}$ video$_{\!}$ editing$_{\!}$ and$_{\!}$ self-driving.$_{\!}$ Prevalent solu- tions$_{\!}$~\cite{jampani2017video,perazzi2017learning,hu2018videomatch,chen2018blazingly,lu2020video,miao2020memory,meinhardt2020make,cheng2021rethinking,oh2019video,yang2020collaborative,caelles2017one,maninis2018video,yang2021associating,bhat2020learning,robinson2020learning,oh2018fast,voigtlaender2019feelvos,wu2020memory,duke2021sstvos,xie2021efficient,liang2020video,hu2022switch,wang2018semi,wang2020paying}$_{\!}$ are$_{\!}$ built$_{\!}$ upon$_{\!}$  \textit{fully$_{\!}$ supervised}$_{\!}$ learning$_{\!}$ techni- ques,$_{\!}$ costing$_{\!}$ intensive$_{\!}$ labeling$_{\!}$ efforts. In$_{\!}$ contrast,$_{\!}$ we$_{\!}$ aim to$_{\!}$~learn$_{\!}$ VOS$_{\!}$ from$_{\!}$ \textit{unlabeled}$_{\!}$ videos$_{\!}$ ---$_{\!}$ \textit{self-supervised}$_{\!}$ VOS.$_{\!}$

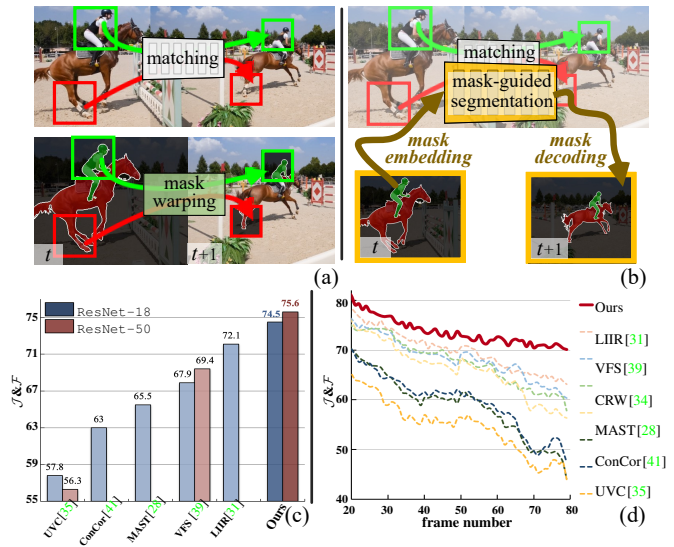
\begin{figure}[t]
\vspace{-6pt}
\centering
      \includegraphics[width=0.99\linewidth]{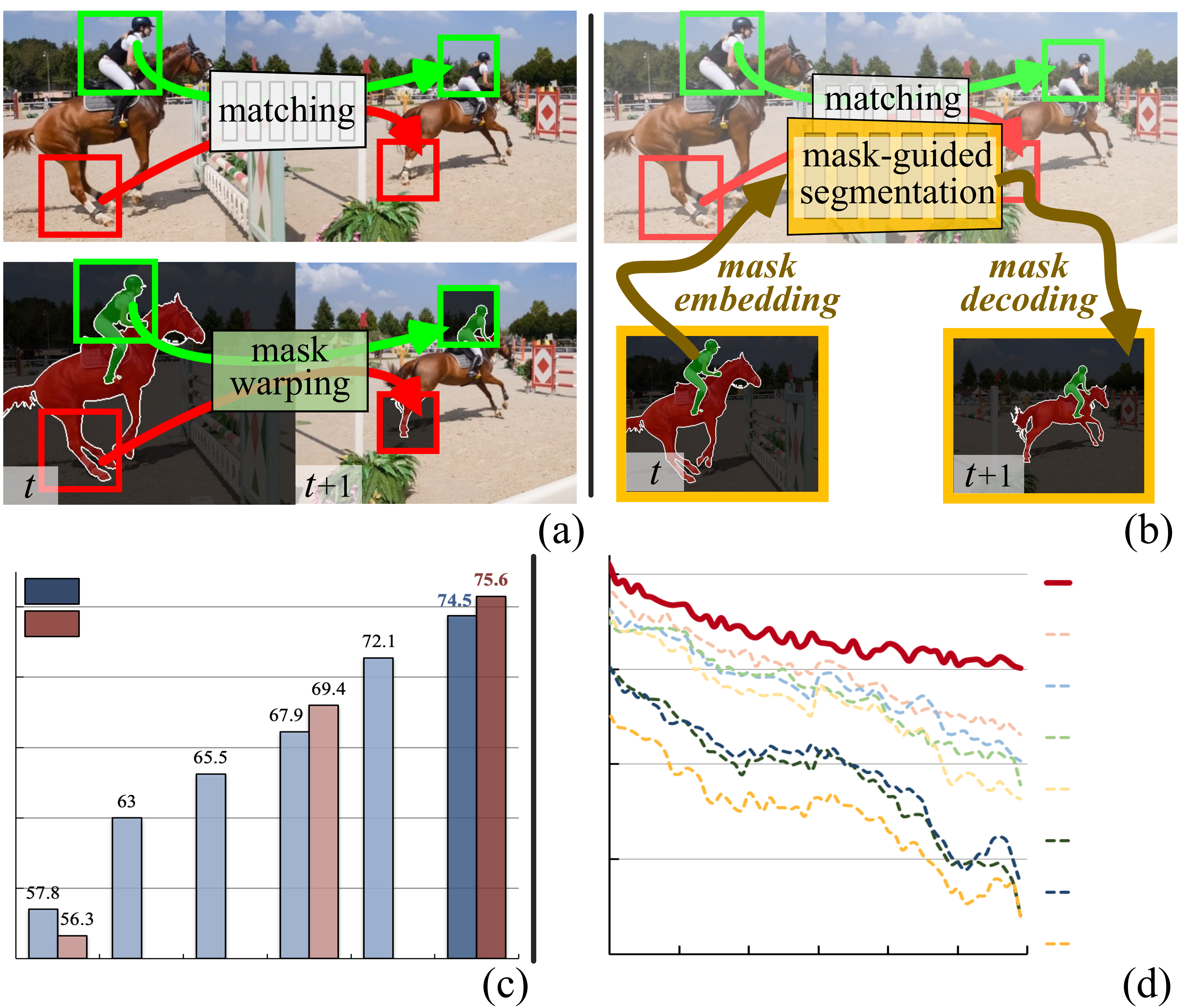}
      \put(-217.9,80){\scalebox{.52}{\texttt{ResNet-18}}}
      \put(-217.9,73.3){\scalebox{.52}{\texttt{ResNet-50}}}
      \put(-199.3,-6.0){\scalebox{.40}{\rotatebox{50}{\textbf{MAST}}}}
      \put(-192.6,3.5){\scalebox{.50}{\rotatebox{50}{\cite{lai2020mast}}}}
      \put(-181.2,-4.5){\scalebox{.40}{\rotatebox{50}{\textbf{VFS}}}}
      \put(-176.3,2.6){\scalebox{.50}{\rotatebox{50}{\cite{xu2021rethinking}}}}
      \put(-167.7,-4.1){\scalebox{.40}{\rotatebox{50}{\textbf{LIIR}}}}
      \put(-163.0,2.6){\scalebox{.50}{\rotatebox{50}{\cite{li2022locality}}}}
      \put(-230.8,-4.5){\scalebox{.40}{\rotatebox{50}{\textbf{UVC}}}}
      \put(-225.9,2.6){\scalebox{.50}{\rotatebox{50}{\cite{li2019joint}}}}
      \put(-217.5,-7.5){\scalebox{.40}{\rotatebox{50}{\textbf{ConCor}}}}
      \put(-209.1,3.6){\scalebox{.50}{\rotatebox{50}{\cite{wang2021contrastive}}}}
      \put(-146.9,-1.5){\scalebox{.50}{\rotatebox{60}{\textbf{Ours}}}}
      \put(-237.2,9.3){\scalebox{.45}{\textbf{55}}}
      \put(-237.2,22.7){\scalebox{.45}{\textbf{59}}}
      \put(-237.2,36.6){\scalebox{.45}{\textbf{63}}}
      \put(-237.2,50.6){\scalebox{.45}{\textbf{67}}}
      \put(-237.2,64.3){\scalebox{.45}{\textbf{71}}}
      \put(-237.2,78.2){\scalebox{.45}{\textbf{75}}}
      \put(-242.6,45){\scalebox{.50}{\rotatebox{90}{\textbf{$\mathcal{J}$\&$\mathcal{F}$}}}}
      \put(-117.3,5.0){\scalebox{.45}{\textbf{20}}}
      \put(-103.0,5.0){\scalebox{.45}{\textbf{30}}}
      \put(-89.0,5.0){\scalebox{.45}{\textbf{40}}}
      \put(-75.0,5.0){\scalebox{.45}{\textbf{50}}}
      \put(-61.0,5.0){\scalebox{.45}{\textbf{60}}}
      \put(-47.0,5.0){\scalebox{.45}{\textbf{70}}}
      \put(-33.0,5.0){\scalebox{.45}{\textbf{80}}}
      \put(-87.4,0){\scalebox{.55}{\textbf{frame number}}}
      \put(-120,9.3){\scalebox{.45}{\textbf{40}}}
      \put(-120,27.8){\scalebox{.45}{\textbf{50}}}
      \put(-120,46.8){\scalebox{.45}{\textbf{60}}}
      \put(-120,65.4){\scalebox{.45}{\textbf{70}}}
      \put(-120,83.5){\scalebox{.45}{\textbf{80}}}
      \put(-124.7,45){\scalebox{.50}{\rotatebox{90}{\textbf{$\mathcal{J}$\&$\mathcal{F}$}}}}
      \put(-22.0,11.6){\scalebox{.55}{UVC\!~\cite{li2019joint}}}
      \put(-22.0,23.2){\scalebox{.55}{ConCor\!~\cite{wang2021contrastive}}}
      \put(-22.0,34.8){\scalebox{.55}{MAST\!~\cite{lai2020mast}}}
      \put(-22.0,46.2){\scalebox{.55}{CRW\!~\cite{jabri2020space}}}
      \put(-22.0,58.2){\scalebox{.55}{VFS\!~\cite{xu2021rethinking}}}
      \put(-22.0,69.5){\scalebox{.55}{LIIR\!~\cite{li2022locality}}}
      \put(-22.0,81.4){\scalebox{.55}{Ours}}
\vspace{-8pt}
	\captionsetup{font=small}
	\caption{\small{$_{\!}$(a)$_{\!}$ Correspondence$_{\!}$ learning$_{\!}$ based$_{\!}$ self-supervised$_{\!}$~VOS, where mask tracking is simply degraded as correspondence matching mask warping. (b)$_{\!}$ We$_{\!}$ achieve$_{\!}$ self-supervised$_{\!}$~VOS$_{\!}$ by$_{\!}$ jointly$_{\!}$ learning$_{\!}$~mask$_{\!}$~embedding$_{\!}$ and$_{\!}$ correspondence$_{\!}$ matching.$_{\!}$ Our algorithm explicitly embeds masks for target object modeling, hence enabling mask-guided segmentation.  (c)$_{\!}$ Performance$_{\!}$ comparison and (d)$_{\!}$ Performance$_{\!}$ over$_{\!}$ time, reported$_{\!}$ on$_{\!}$ DAVIS$_{17}$\!~\cite{perazzi2016benchmark} \texttt{val}. }}
\label{fig:overview}
\vspace{-15pt}
\end{figure}

Due to the absence of mask annotation during training, existing studies typically degrade$_{\!}$ such$_{\!}$ self-supervised$_{\!}$ yet$_{\!}$ \textit{mask-guided$_{\!}$ segmentation}$_{\!}$ task$_{\!}$~as$_{\!}$ a$_{\!}$ combo$_{\!}$ of$_{\!}$~\textit{unsupervised  correspondence$_{\!}$ learning}$_{\!}$ and$_{\!}$ correspondence$_{\!}$ based,$_{\!}$~\textit{non- learnable$_{\!}$ mask$_{\!}$ warping}$_{\!}$ (\textit{cf.}$_{\!}$~Fig.$_{\!}$~\ref{fig:overview}(a)).$_{\!}$ They$_{\!}$ first$_{\!}$ learn$_{\!}$~pixel-
 /patch-wise$_{\!}$ matching$_{\!}$ (\ie, cross-frame$_{\!}$ correspondence)$_{\!}$ by exploring$_{\!}$ the$_{\!}$ inherent$_{\!}$ continuity$_{\!}$ in$_{\!}$ raw$_{\!}$ videos$_{\!}$ as$_{\!}$ free$_{\!}$~super- visory signals, in the form of \textbf{i)} a \textit{photometric} \textit{reconstruction} problem  where each pixel in a target frame is desired to be
recovered by$_{\!}$ \textit{copying}$_{\!}$ relevant$_{\!}$ pixels$_{\!}$ in$_{\!}$ reference$_{\!}$ frame(s) \cite{vondrick2018tracking,lai2019self,lai2020mast,kim2020rpm,yang2021self,li2022locality}; \textbf{ii)}  a$_{\!}$~\textit{cycle-consistency}$_{\!}$ task that$_{\!}$ enforces$_{\!}$ matching$_{\!}$ of$_{\!}$ pixels/patches$_{\!}$ after$_{\!}$ forward-backward$_{\!}$ tracking \cite{wang2019learning,wang2019unsupervised,jabri2020space,li2019joint,bian2022learning};$_{\!}$ and \textbf{iii)} a$_{\!}$~\textit{contrastive$_{\!}$ matching}$_{\!}$ scheme that contrasts confi- dent$_{\!}$ correspondences$_{\!}$ against$_{\!}$ unreliable$_{\!}$~ones$_{\!}$ \cite{jeon2021mining,araslanov2021dense,xu2021rethinking,son2022contrastive}.$_{\!}$ Once trained, the dense matching model is used to approach VOS in a cheap way (Fig.$_{\!\!}$~\ref{fig:overview}(a)): the label of a query pixel/patch is simply$_{\!}$
 borrowed from previously segmented ones, according~to their appearance similarity (correspondence score).

$_{\!}$Though$_{\!}$ straightforward,$_{\!}$ these$_{\!}$ correspondence$_{\!}$ based~$_{\!}$``ex- pedient'' solutions come with~two$_{\!}$ severe limitations:$_{\!}$ \textbf{First},$_{\!}$ they learn to match pixels instead of customizing
VOS target -- mask-guided segmentation, leaving a significant~gap between the training goal and task/inference setup. During  training,$_{\!}$ the$_{\!}$ model$_{\!}$ is$_{\!}$ optimized$_{\!}$ purely$_{\!}$ to$_{\!}$ discovery$_{\!}$~reliable,  target-agnostic  visual$_{\!}$ correlations,$_{\!}$ with$_{\!}$ no$_{\!}$ sense$_{\!}$ of$_{\!}$ object-mask$_{\!}$ information.$_{\!}$ Spontaneously,$_{\!}$ during testing/inference,$_{\!}$ the$_{\!}$ model$_{\!}$ struggles$_{\!}$ in$_{\!}$ employing$_{\!}$ first-/prior-frame$_{\!}$ masks$_{\!}$~to
 guide$_{\!}$ the$_{\!}$ prediction$_{\!}$ of$_{\!}$ succeeding$_{\!}$ frames. \textbf{Second},$_{\!}$ from$_{\!}$ the$_{\!}$ view$_{\!}$ of$_{\!}$ mask-tracking,$_{\!}$ existing$_{\!}$  self-supervised$_{\!}$~solutions,$_{\!}$ in
 essence,$_{\!}$ adopt$_{\!}$ an$_{\!}$ obsolete,$_{\!}$ matching-/flow-based mask$_{\!}$ pro-
 pagation$_{\!}$ strategy$_{\!}$~\cite{badrinarayanan2010label,jain2014supervoxel,avinash2014seamseg,wang2017selective,wang2017super}.$_{\!}$ As$_{\!}$ discussed$_{\!}$ even$_{\!}$ before$_{\!}$ the$_{\!}$ deep learning era$_{\!}$~\cite{faktor2014video,fan2015jumpcut,wang2015saliency}, such a strategy is sub-optimal. Specifi- cally, without$_{\!}$ modeling$_{\!}$ the$_{\!}$ target$_{\!}$ objects,$_{\!}$ flow-based$_{\!}$ mask warping$_{\!}$ is$_{\!}$ sensitive$_{\!}$ to$_{\!}$ outliers,$_{\!}$ resulting$_{\!}$ in$_{\!}$ error$_{\!}$~accumula-
tion$_{\!}$ over$_{\!}$ time$_{\!}$~\cite{wang2021survey}.$_{\!}$ Subject~to$_{\!}$ the$_{\!}$ primitive$_{\!}$ matching-and-copy mechanism,$_{\!}$ even$_{\!}$ trivial$_{\!}$ errors$_{\!}$ are$_{\!}$ hard$_{\!}$ to$_{\!}$ be$_{\!}$ corrected,$_{\!}$ and often$_{\!}$ lead$_{\!}$ to$_{\!}$ much$_{\!}$ worse$_{\!}$ results$_{\!}$ caused$_{\!}$ by$_{\!}$ drifts$_{\!}$ or$_{\!}$~occlusions. This is also why current top-leading \textit{fully} \textit{supervised} VOS$_{\!}$ solutions$_{\!}$~\cite{caelles2017one,maninis2018video,yang2021associating,bhat2020learning,robinson2020learning,oh2018fast,hu2018videomatch,chen2018blazingly,oh2019video,voigtlaender2019feelvos,wu2020memory,yang2020collaborative,duke2021sstvos,xie2021efficient,liang2020video}$_{\!}$ largely follow$_{\!}$ a$_{\!}$ \textit{mask$_{\!}$ embedding learning}$_{\!}$ philosophy$_{\!}$ ---$_{\!}$ embedding \textit{frame-mask} \textit{pairs}, instead of only frame images,$_{\!}$ into the segmentation network. With such explicit modeling of the target object, more robust and accurate mask-tracking can be achieved~\cite{wang2021survey,JAS-2022-1528}.

Motivated by the aforementioned discussions, we integrate mask embedding learning and dense correspondence modeling into a compact, end-to-end framework for self-supervised$_{\!}$ VOS$_{\!}$  (\textit{cf.}$_{\!}$~Fig.$_{\!\!}$~\ref{fig:overview}(b)).$_{\!}$ This$_{\!}$ allows$_{\!}$ us$_{\!}$ to$_{\!}$ inject$_{\!}$~the mask-tracking nature of the task  into$_{\!}$ the$_{\!}$ very$_{\!}$ heart$_{\!}$ of$_{\!}$ our$_{\!}$ algorithm and model training.$_{\!}$ However,$_{\!}$ bringing$_{\!}$ the$_{\!}$ idea$_{\!}$~of mask embedding into self-supervised$_{\!}$ VOS$_{\!}$ is$_{\!}$ not$_{\!}$ trivial,$_{\!}$ due\\
\noindent to the lack of mask annotation. We therefore achieve mask embedding$_{\!}$ learning$_{\!}$ in$_{\!}$ a$_{\!}$ \textit{self-taught}$_{\!}$ manner.$_{\!}$ Concretely,$_{\!}$~our model is trained by alternating between$_{\!}$ \textbf{i)} space-time$_{\!}$ pixel$_{\!}$ clustering, and \textbf{ii)} mask-embedded segmentation learning. Pixel clustering is to automatically discover spatiotempo-
rally$_{\!}$ coherent$_{\!}$ object(-like)$_{\!}$ regions$_{\!}$ from$_{\!}$ raw$_{\!}$ videos.$_{\!}$ By$_{\!}$ uti- lizing such pixel-level video partitions as pseudo ground- truths$_{\!}$ of$_{\!}$ target$_{\!}$ objects,$_{\!}$ our$_{\!}$ model$_{\!}$ can$_{\!}$ learn$_{\!}$ how$_{\!}$ to extract target-specific context from frame-mask  pairs, and how to~leverage such$_{\!}$ high-level$_{\!}$ context$_{\!}$ to$_{\!}$ predict$_{\!}$ the$_{\!}$ next-frame mask. At the same time, such self-taught mask embedding scheme is consolidated by self-supervised dense correspondence learning. This allows our model to learn transferable, locally discriminative representations by making full use of the spatiotemporal coherence in~natural videos, and prevent the degenerate solution of the deterministic clustering.

$_{\!}$Our$_{\!}$ approach$_{\!}$ owns$_{\!}$ a$_{\!}$ few$_{\!}$ distinctive$_{\!}$ features:$_{\!}$ \textbf{First},$_{\!}$ it$_{\!}$~has the$_{\!}$ ability$_{\!}$ of$_{\!}$ directly$_{\!}$ learning$_{\!}$ to$_{\!}$  conduct$_{\!}$ mask-guided$_{\!}$~se- quential segmentation; its training~objective~is~completely aligned with the core nature of VOS. \textbf{Second}, by  learning to embed object-masks into mask tracking, target-oriented context can be efficiently mined and explicitly leveraged for object modeling, rather than existing methods merely relying on local appearance~correlations for label ``copying''. Hence our approach can reduce error accumulation (\textit{cf.}$_{\!}$~Fig.$_{\!}$~\ref{fig:overview}(d)) and perform more robust when the latent correspondences are ambiguous,$_{\!}$ \eg,$_{\!}$~deformation, occlusion or one-to-many matches. \textbf{Third}, our mask embedding strategy endows our self-supervised framework with the potential of being empowered by more advanced VOS model designs developed in the fully-supervised learning setting.

Through embracing the powerful idea of mask embedding learning as well as inheriting the merits of correspon- dence$_{\!}$ learning,$_{\!}$ our$_{\!}$ approach$_{\!}$ favorably$_{\!}$ outperforms$_{\!}$~state-of-\\
\noindent the-art competitors, \ie, \textbf{3.2}\%, \textbf{2.5}\%, and \textbf{2.2}\%  mIoU gains on DAVIS$_{17}$$_{\!}$~\cite{perazzi2016benchmark} \texttt{val}, DAVIS$_{17}$ \texttt{test-dev} and YouTube-VOS$_{\!}$~\cite{xu2018youtube} \texttt{val}, respectively. In addition to narrowing the performance gap between self- and fully-supervised VOS, our approach establishes a tight coupling between them in the aspect of model design. We expect this work can foster the mutual collaboration between these two relevant fields.

\section{Related Work}\label{sec:relatedwork}
\vspace{-1pt}
\noindent\textbf{Fully$_{\!}$ Supervised$_{\!}$ Learning$_{\!}$ for$_{\!\!}$ VOS.$_{\!}$} Given$_{\!}$ a$_{\!}$ target$_{\!}$ object$_{\!}$ outlined$_{\!}$ in$_{\!}$ the$_{\!}$ first frame, VOS aims to precisely extract this object from the rest frames. Fully supervised deep learning$_{\!}$ based solutions have became the mainstream
in this$_{\!}$ field,$_{\!}$ and$_{\!}$ can$_{\!}$ be$_{\!}$ broadly$_{\!}$ grouped$_{\!}$ into three categories$_{\!}$~\cite{wang2021survey}: \textit{online finetuning} based$_{\!}$~\cite{bhat2020learning,robinson2020learning} (\ie, training a~segmentation network separately on each test-time given object), \textit{propagation} based~\cite{oh2018fast,caelles2017one,maninis2018video} (\ie, using the latest mask to infer the upcoming frame mask), and \textit{matching} based$_{\!}$~\cite{hu2018videomatch,chen2018blazingly,oh2019video,voigtlaender2019feelvos,yang2020collaborative,duke2021sstvos} (\ie, classifying pixels according to their similarities to the target object). Despite assorted motivations and technique details, almost all the top-leading approaches are built upon a common principle --~embedding paired frame and mask, \eg, $(I_r, Y_r)$, into the segmentation network $\mathcal{S}$:
\vspace{-1pt}
\begin{equation}\small\label{eq:el}
	Y_q = \mathcal{S}(I_q, \{\mathcal{V}(I_r, Y_r)\}_r),
\end{equation}
where $Y_q$ is the mask predicted for a given query frame $I_q$; the function $\mathcal{V}$ learns to deduce target-specific context from\\
\noindent  the$_{\!}$ reference$_{\!}$ $(I_r, Y_r)$;$_{\!}$ $\{(I_r,_{\!} Y_r)\}_{r\!}$ can$_{\!}$ be$_{\!}$ the$_{\!}$ initial$_{\!}$ frame-mask$_{\!}$ pair$_{\!}$~\cite{chen2018blazingly,hu2018videomatch,voigtlaender2019feelvos,yang2020collaborative,oh2018fast},$_{\!}$ and/or$_{\!}$ paired$_{\!}$  historical$_{\!}$ frames$_{\!}$ and mask$_{\!}$ predictions$_{\!}$~\cite{zhang2020transductive,oh2019video,lu2020video,seong2020kernelized,hu2021learning,cheng2021rethinking,duke2021sstvos}.$_{\!}$
Then$_{\!}$ the$_{\!}$ target-aware$_{\!}$ context$_{\!}$ is$_{\!}$ used$_{\!}$ to$_{\!}$ guide  the$_{\!}$ segmentation$_{\!}$ (mask$_{\!}$~de- coding)$_{\!}$ of$_{\!}$ the$_{\!}$ query$_{\!}$ $I_q$.$_{\!}$ For$_{\!}$~instance,$_{\!}$ \cite{chen2018blazingly,hu2018videomatch,voigtlaender2019feelvos,yang2020collaborative,hu2023suppressing}$_{\!}$ store    foreground$_{\!}$ and$_{\!}$ background$_{\!}$ representations$_{\!}$ and$_{\!}$ classify$_{\!}$ pi- xels by nearest neighbor retrieval or feature decoding. Some others directly project reference frame and mask into a joint (space-time) embedding space, which is subsequently used for mask propagation$_{\!}$~\cite{oh2018fast} or feature matching$_{\!}$~\cite{oh2019video,lu2020video,seong2020kernelized,hu2021learning,cheng2021rethinking,duke2021sstvos}. A few recent methods \cite{bhat2020learning,mao2021joint} treat mask-derived object representation as the target of a few-shot learner \cite{bhat2020learning} or a prior for joint inductive and transductive learning \cite{mao2021joint}.

$_{\!}$Inspired$_{\!}$ by$_{\!}$ these$_{\!}$ achievements,$_{\!}$ for$_{\!}$ the$_{\!}$ first$_{\!}$ time,$_{\!}$ we$_{\!}$~ex- ploit$_{\!}$ the$_{\!}$ idea$_{\!}$ of$_{\!}$ mask$_{\!}$ embedding$_{\!}$ in$_{\!}$ the$_{\!}$ self-supervised$_{\!\!}$~VOS$_{\!}$ setting.$_{\!}$ We$_{\!}$ achieve$_{\!}$ this$_{\!}$ through$_{\!}$ automatic space-time~clus- tering$_{\!}$ and$_{\!}$ using$_{\!}$  deterministic$_{\!}$ cluster$_{\!}$ assignments$_{\!}$ as$_{\!}$ pseudo$_{\!}$ groundtruths to supervise the learning of mask embedding and decoding, without the aid of manual labels. In this way, our self-supervised model is capable of explicitly and comprehensively describing the target object, hence achieving more robust and accurate, target-oriented segmentation.

\noindent\textbf{Self-supervised$_{\!}$ Learning$_{\!}$ for$_{\!}$ VOS.$_{\!}$} Learning$_{\!}$ VOS$_{\!}$ in$_{\!}$ a$_{\!}$ self-supervised$_{\!}$ manner$_{\!}$ is$_{\!}$ appealing,$_{\!}$ as$_{\!}$ it$_{\!}$ eliminates$_{\!}$  the$_{\!}$ heavy annotation budget required by the fully supervised algori- thms. Due to the absence of mask annotation, existing self-supervised$_{\!}$ methods$_{\!}$ take$_{\!}$ an$_{\!}$ \textit{expedient}$_{\!}$ solution:$_{\!}$ they$_{\!}$ learn$_{\!}$ to$_{\!}$ find$_{\!}$ correspondence$_{\!}$ between$_{\!}$ two frames, instead of learning mask-guided segmentation. During inference, the first-frame mask is directly$_{\!}$ copied$_{\!}$ to$_{\!}$ the$_{\!}$ rest$_{\!}$ frames$_{\!}$ based$_{\!}$ on$_{\!}$ cross-frame correspondence. Specifically, given two frames $I_r$ and$_{\!}$ $I_q$,$_{\!}$ their$_{\!}$ dense$_{\!}$ representations$_{\!}$ $\bm{I}_r, \bm{I}_{q\!}\!\in_{\!}\!\mathbb{R}^{H_{\!}W_{\!}\times_{\!}D\!\!}$~are$_{\!}$ first extracted$_{\!}$ by$_{\!}$ a$_{\!}$ shallow$_{\!}$ neural$_{\!}$ encoder$_{\!}$ $\mathcal{E}$ (typically \texttt{ResNet- 18}$_{\!}$~\cite{he2016deep}), and their pairwise affinity matrix can be computed as:
\vspace{-3pt}
\begin{equation}\small\label{eq:correspondence}
	{A}_r^q=\texttt{softmax}(\bm{I}_r \bm{I}_q^\top)~~\in[0,1]^{HW\!\times\!HW},
\vspace{-3pt}
\end{equation}
where$_{\!}$ $\texttt{softmax}$$_{\!}$ is$_{\!}$ row-wise.$_{\!}$ The$_{\!}$ resultant$_{\!}$ affinity$_{\!}$ $A_{r\!}^q$~gives$_{\!}$ the$_{\!}$ strength$_{\!}$ of$_{\!}$ all$_{\!}$ the$_{\!}$ pixel$_{\!}$ pairwise$_{\!}$ correspondence$_{\!}$ between $\bm{I}_r$ and $\bm{I}_q$. One$_{\!}$ main$_{\!}$ benefit$_{\!}$ is that, once$_{\!}$ $\mathcal{E}$$_{\!}$ is$_{\!}$ trained,$_{\!}$ it$_{\!}$ can$_{\!}$ be$_{\!}$ used$_{\!}$ to$_{\!}$ estimate$_{\!}$ cross-frame$_{\!}$ correspondence;$_{\!}$ then$_{\!}$ VOS is$_{\!}$ approached$_{\!}$ by$_{\!}$ warping$_{\!}$ the$_{\!}$  mask$_{\!}$ $Y_{r\!}$ of$_{\!}$ a$_{\!}$ reference$_{\!}$ frame$_{\!}$ $I_{r\!}$ to$_{\!}$~the$_{\!}$ query$_{\!}$ frame ${I}_{q\!}$~based$_{\!}$ on:$_{\!}$ $Y_q\!=\!A_r^{q\top\!}Y_r$.$_{\!}$
Thus$_{\!}$ the$_{\!}$ central problem is to design a surrogate  task~to supervise $\mathcal{E}$ to estimate reliable intra-frame affinity ${A}_r^q$. Basically, three~types of surrogate tasks are developed, yet all exploit the correlation among frames: \textbf{i)} \textit{Photometric$_{\!}$ reconstruction}$_{\!}$~\cite{vondrick2018tracking,lai2019self,lai2020mast,wang2021contrastive,jeon2021mining,araslanov2021dense,bhat2020learning}.$_{\!}$ Here$_{\!}$ the$_{\!}$ affinity$_{\!}$ matrix$_{\!}$ $A_r^{q\!}$ is$_{\!}$ estimated to reconstruct the query frame: $\tilde{I}_q\!=\!A_r^{q\top\!}I_r$, invoking a photo- metric reconstruction$_{\!}$ objective:  $\mathcal{L}_{\text{Re}}\!=\!\|I_{q\!}-_{\!}\tilde{I}_q\|^2$; \textbf{ii)} \textit{Cycle- consistency tracking}$_{\!}$~\cite{wang2019unsupervised,wang2019learning,li2019joint,lu2020learning,jabri2020space,zhao2021modelling,li2020content}.$_{\!}$ The$_{\!}$ affinity$_{\!}$ $A_r^q$$_{\!}$ is$_{\!}$ used\\
\noindent to$_{\!}$ guide$_{\!}$ a$_{\!}$ cycle of forward and backward tracking, leading$_{\!}$ to$_{\!}$ a$_{\!}$ cycle-consistency$_{\!}$ loss:$_{\!}$  $\mathcal{L}_{\text{Cyc}}\!=\!\|A_r^qA_q^r\!-\!\mathbbm{1}\|^2$,$_{\!}$ where$_{\!}$ $\mathbbm{1}$$_{\!}$~is$_{\!}$\\
\noindent an$_{\!}$ identity$_{\!}$~matrix$_{\!}$  with$_{\!}$ proper$_{\!}$ size;$_{\!}$ and$_{\!}$ \textbf{iii)}$_{\!}$ \textit{Contrastive$_{\!}$~ma-} \textit{tching}$_{\!}$~\cite{jeon2021mining,araslanov2021dense,xu2021rethinking}.$_{\!}$ Drawn$_{\!}$ inspiration  from unsupervised contrastive  learning$_{\!}$~\cite{oord2018representation,chen2020simple},$_{\!}$ temporal$_{\!}$ correspondence$_{\!}$ learning is achieved by contrasting the affinity between positive, matched pixel pairs $(i,i^+)$, against the affinity between ne- gative,  unrelated ones $(i,i^-)$: $\mathcal{L}_{\text{Con}}\!=\!-\log(\exp(A(i,i^+))/$ $\sum_{i^-\!}\exp(A(i,i^-)))$.$_{\!}$ The$_{\!}$ positive$_{\!}$ pairs$_{\!}$ are$_{\!}$ often$_{\!}$ pre-defined as spatiotemporally adjacent pixels, so as to capture the~co- herence$_{\!}$ residing$_{\!}$ videos$_{\!}$~\cite{jeon2021mining,araslanov2021dense},$_{\!}$ while$_{\!}$ \cite{xu2021rethinking}$_{\!}$ shows$_{\!}$ that$_{\!}$~fine- grained correspondence can be captured by directly contrasting frame samples. For long-range matching, multiple reference frames are considered in practice$_{\!}$~\cite{lai2020mast,jabri2020space,jeon2021mining,araslanov2021dense}. 

\begin{figure*}[t]
\centering
      \includegraphics[width=0.99\linewidth]{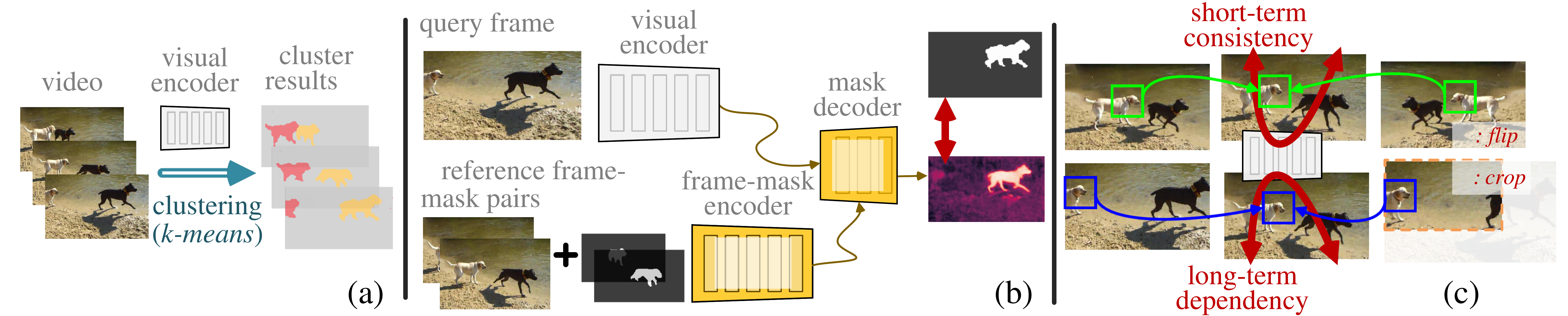}
      \put(-466,15.3){\scalebox{0.8}{$\mathcal{I}$}}
      \put(-384,71.3){\scalebox{0.8}{$\bm{S}^*$}}
      \put(-315,88.5){\scalebox{0.8}{$I_q$}}
      \put(-321,33.8){\scalebox{0.8}{$(I_{r_n},Y_{r_n})_n$}}
      \put(-211.5,88.8){\scalebox{0.8}{$Y_{q}$}}
      \put(-210,23.5){\scalebox{0.8}{$\hat{Y}_{q}$}}
      \put(-153,82.5){\scalebox{0.8}{$I_{t+1}$}}
      \put(-151,13.5){\scalebox{0.8}{$I_{t'}$}}
      \put(-72.6,47.3){\scalebox{1.0}{$I_{t}$}}
      \put(-40,84.8){\scalebox{0.8}{$\Phi (I_{t+1})$}}
      \put(-39,17.5){\scalebox{0.8}{$\Phi (I_{t'})$}}
      \put(-35.2,53.2){\scalebox{0.92}{\textcolor[RGB]{183,0,24}{$\Phi$}}}
      \put(-35.2,40.5){\scalebox{0.92}{\textcolor[RGB]{183,0,24}{$\Phi$}}}
      \put(-431.8,55.4){\scalebox{0.9}{$\mathcal{E}$}}
      \put(-287.2,63.9){\scalebox{0.9}{$\mathcal{E}$}}
      \put(-258.2,13.4){\scalebox{0.9}{$\mathcal{V}$}}
      \put(-225.3,44.1){\scalebox{0.9}{$\mathcal{D}$}}
      \put(-91.0,47.0){\scalebox{0.9}{$\mathcal{E}$}}
      \put(-190.1,57.2){\color{myred}{\scalebox{0.9}{$\mathcal{L}_\text{Seg}$}}}
      \put(-74,88.3){\color{myred}{\scalebox{0.9}{$\mathcal{L}_\text{Short}$}}}
      \put(-75,7.3){\color{myred}{\scalebox{0.9}{$\mathcal{L}_\text{Long}$}}}
\vspace{-5pt}
\caption{$_{\!}$Our$_{\!}$ self-supervised$_{\!}$ VOS$_{\!}$ framework:$_{\!}$ \textbf{(a-b)}$_{\!}$ space-time pixel clustering$_{\!}$ based$_{\!}$ mask$_{\!}$ embedding$_{\!}$ learning (\S\ref{sec:framework}) for the whole net- work (including $\mathcal{E}$, $\mathcal{V}$, and $\mathcal{D}$), and \textbf{(c)} short- and long-term correspondence learning (\S\ref{sec:correspondence}) for the visual encoder $\mathcal{E}$ only. }
\label{fig:framework}
\vspace{-10pt}
\end{figure*}
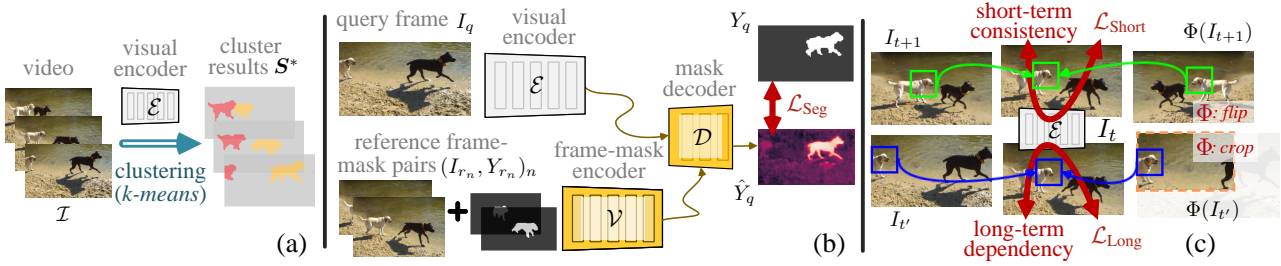

Our$_{\!}$ algorithm$_{\!}$ is$_{\!}$ fundamentally$_{\!}$ different$_{\!}$ from$_{\!}$ existing$_{\!}$ self-supervised$_{\!}$ VOS$_{\!}$~solutions.$_{\!}$ Through$_{\!}$ self-taught$_{\!}$ mask$_{\!}$ embedding$_{\!}$ learning,$_{\!}$ our$_{\!}$ method begets {mask-guided~seg- mentation}. Thus the nature of VOS is captured by our network architecture and training, rather than existing methods treating the task as a derivant of unsupervised correspondence learning. Further, our method is principled; correspondence learning can be seamlessly incorporated for regularizing representation learning, while the concomitant shortcomings, \eg, no sense of target-specific information, error accumulation over time, and misalignment between training and inference modes, are naturally alleviated. 

\section{Methodology} \label{sec:methodology}
\vspace{-1pt}
At a high level, our self-supervised VOS solution~jointly learns$_{\!}$ mask$_{\!}$ embedding$_{\!}$ and$_{\!}$ visual$_{\!}$ correspondence$_{\!}$ from$_{\!}$~raw videos. It absorbs~the powerful$_{\!}$ idea$_{\!}$ of$_{\!}$ mask-embedded$_{\!}$ seg- mentation$_{\!}$ ($_{\!}$\textit{cf}.$_{\!}$~Eq.$_{\!}$~\ref{eq:el})$_{\!}$ in$_{\!}$ fully$_{\!}$ supervised$_{\!}$ VOS; meanwhile, it inherits the merits$_{\!}$ of$_{\!}$ existing$_{\!}$ unsupervised$_{\!}$ correspondence$_{\!}$ based$_{\!}$ regime$_{\!}$ ($_{\!}$\textit{cf}.$_{\!}$~Eq.$_{\!}$~\ref{eq:correspondence})$_{\!}$ in$_{\!}$~learning$_{\!}$ generic,$_{\!}$ dense$_{\!}$ features.$_{\!}$ As a result,$_{\!}$~our solution can be formulated as ($_{\!}$\textit{cf}.$_{\!}$~Fig.$_{\!}$~\ref{fig:framework}):
\vspace{-3pt}
\begin{equation}
\small\label{eq:overview}
Y_q =
    \tikzmarknode{qp1}{\highlight{myy}{\color{black}$\!\mathcal{D}\big(\!\!$}}
    \tikzmarknode{qp2}{\highlight{myg}{\color{black}${\!\!\mathcal{E}(I_q),}\!$}}
    \tikzmarknode{qp3}{\highlight{myy}{\color{black}$\!\!\{\mathcal{V}([I_{r_n}, \!\!$}}
    \tikzmarknode{qp4}{\highlight{myy}{\color{black}$\!\!Y_{r_n}])\}_n\big)\!\!$}}
\end{equation}
\vspace*{0.6\baselineskip}
\begin{tikzpicture}[overlay,remember picture,>=stealth,nodes={align=left,inner ysep=1pt},<-]
\path (qp2.north) ++ (-3.27,-1.8em) node[anchor=north west,color=myg2] (tj1text){
   $\substack{
      \text{\scriptsize self-supervised dense}\\
      \text{\scriptsize correspondence learning \S\ref{sec:correspondence}}
   }$
};
\draw [color=myg2](qp2.south) |- ([xshift=0.3ex,color=myg]tj1text.south west);
\path (qp3.north) ++ (0,-1.8em) node[anchor=north west,color=myy2] (tjtext){
   $\substack{
      \text{\scriptsize self-supervised}\\
      \text{\scriptsize mask embedding learning \S\ref{sec:framework}}
   }$
};
\draw [color=myy2](qp3.south) |- ([xshift=-1.0ex,color=myy]tjtext.south east);
\end{tikzpicture}

\noindent  where$_{\!}$ $[_{\!\!}~\cdot_{\!\!}~]$$_{\!}$ stands$_{\!}$ for$_{\!}$ concatenation.$_{\!}$ Basically,$_{\!}$ our$_{\!}$ model$_{\!}$ uti- lizes$_{\!}$ a$_{\!}$~set$_{\!}$ of$_{\!}$ reference$_{\!}$ frame-mask$_{\!}$ pairs,$_{\!}$ \ie,$_{\!}$ $(I_{r_n\!},_{\!} Y_{r_n\!})_n$,~to predict/decode the mask of each query frame $I_q$, learnt in a self-supervised manner. Our model has three core parts:
\begin{itemize}[leftmargin=*]
	\setlength{\itemsep}{3pt}
	\setlength{\parsep}{1pt}
	\setlength{\parskip}{1pt}
	\setlength{\leftmargin}{-10pt}
	\vspace{-6pt}
	\item \textbf{\textit{Visual Encoder}} $\mathcal{E}$, which maps each query frame $I_q$ into a dense representation$_{\!}$ tensor:$_{\!}$ $\bm{I}_{q\!}\!=\!\mathcal{E}(I_q)\!\in\!\mathbb{R}^{HW\!\times\!D\!}$.$_{\!}$ We instantiate $\mathcal{E}$ as \texttt{ResNet-18} or \texttt{ResNet-50}.
	\vspace{-4pt}
	\item \textbf{\textit{Frame-Mask Encoder}} $\mathcal{V}$ for mask embedding. It takes a pair of a reference frame $I_r$ and corresponding mask $Y_r$ as inputs, and extracts target-specific context, \ie, $\bm{V}_{r\!}\!=\!\mathcal{V}([I_r, Y_r])\!\in\!\mathbb{R}^{HW\!\times\!D'\!}$, to guide the segmentation/mask decoding of ${I}_{q}$. $\mathcal{V}$ has a similar network architecture with $\mathcal{E}$, but the input and output dimensionality are different and the network weights are unshared.
\vspace{-4pt}
	\item \textbf{\textit{Mask Decoder}} $\mathcal{D}$, which is a small CNN for mask~deco- ding.$_{\!}$ With$_{\!}$ the$_{\!}$ help$_{\!}$ of$_{\!}$ target-rich$_{\!}$  context $\{\bm{V}_{r_n\!}\}_{n\!}$ collected from$_{\!}$ a$_{\!}$ set$_{\!}$ of$_{\!}$ reference$_{\!}$ frame-mask$_{\!}$ pairs$_{\!}$ $\{(I_{r_n}, Y_{r_n})\}_{n}$,$_{\!}$ $\mathcal{D}_{\!}$ makes robust prediction, \ie, $Y_q$, for the query frame ${I}_{q}$.
\end{itemize}
\vspace{-6pt}

\noindent As for training, to mitigate the dilemma caused by the~ab- sence$_{\!}$ of$_{\!}$ true$_{\!}$ labels$_{\!}$ of$_{\!}$ $\{Y_{r_n}\}_{n\!}$ and$_{\!}$ $Y_q$,$_{\!}$ we$_{\!}$ conduct$_{\!}$ unsuper- vised space-time pixel clustering for automatic mask crea- tion and train the whole network, including~$\mathcal{E}$, $\mathcal{V}$, and $\mathcal{D}$, for\\
\noindent mask$_{\!}$ embedding$_{\!}$ and$_{\!}$ decoding$_{\!}$ (\S\ref{sec:framework}).$_{\!}$ Moreover,$_{\!}$ unsupervised contrastive$_{\!}$ correspondence learning$_{\!}$ (\S\ref{sec:correspondence}) is introduced to boost dense visual representation learning of $\mathcal{E}$.

\vspace{-0pt}
\subsection{Self-supervised Mask Embedding Learning} \label{sec:framework}\vspace{-2pt}
For self-supervised mask embedding learning, we alternatively perform two steps: \textbf{Step$_{\!\!}$ 1}:$_{\!}$ clustering$_{\!}$ of$_{\!}$ video$_{\!}$ pixels$_{\!}$ on$_{\!}$ the$_{\!}$ visual$_{\!}$ feature$_{\!}$ space$_{\!}$ $\mathcal{E}$$_{\!}$ so$_{\!}$ as$_{\!}$~to$_{\!}$ generate spatiotemporally compact segments; and \textbf{Step$_{\!}$ 2}: the space-time cluster\\
\noindent assignments serve as pseudo masks to supervise our whole network$_{\!}$ (including$_{\!}$ $\mathcal{E}$,$_{\!}$ $\mathcal{V}$,$_{\!}$ and$_{\!}$ $\mathcal{D}$),$_{\!}$ which$_{\!}$ learns$_{\!}$ VOS$_{\!}$ as mask- embedded$_{\!}$ sequential$_{\!}$ segmentation.$_{\!}$ After$_{\!}$ that,$_{\!}$ the$_{\!}$ improved$_{\!}$ visual representation $\mathcal{E}$ will in turn facilitate clustering.

\noindent\textbf{Step$_{\!\!}$ 1:$_{\!}$ Space-time$_{\!}$ Clustering.}$_{\!}$ The goal of this step is to partition each training video $\mathcal{I}$ into $M$  space-time consistent segments (see~Fig.$_{\!}$~\ref{fig:framework}(a)). For each pixel $i\!\in\!\mathcal{I}$, let $\bm{i}\!\in\!\mathbb{R}^{D\!}$ denote its visual embedding (extracted from the visual encoder $\mathcal{E}$), and $\bm{s}_i\!\in\!\{0,1\}^{M\!}$ its one-hot cluster assignment vector. Clustering of all the pixels in $\mathcal{I}$ into $M$ clusters can be achieved by solving the following optimization problem:
\vspace{-3pt}
\begin{equation}\small
\label{eq:kmeans}
   \min_{\bm{C}, \bm{S}} \sum\nolimits_{i\in\mathcal{I}} \| \bm{i} - \bm{C}\bm{s}_i \|, ~~~~ \textit{s.t.}~~\bm{s}_i\in\{0,1\}^M,~~~\bm{1}^\top \bm{s}_i=1.
\vspace{-3pt}
\end{equation}
Here $\bm{C}\!=\![\bm{c}_1, \ldots\!, \bm{c}_M]\!\in\!\mathbb{R}^{D\times M\!}$ is the cluster centroid matrix, where $\bm{c}_m\!\in\!\mathbb{R}^{D\!}$~refers to the centroid of $m$-\textit{th} cluster, and $\bm{S}\!=\![\bm{s}_i]_i$ stores the cluster assignments of all the pixels in $\mathcal{I}$. $\bm{1}$ is a $M$-dimensional all-one vector.  While many clustering~methods have been designed to solve Eq.$_{\!}$~\ref{eq:kmeans}, for simplicity,  we use the most~classic one$_{\!}$ --$_{\!}$ $k$-means,$_{\!}$ which$_{\!}$ finds the optimal $\bm{C}^*$ and $\bm{S}^*$ in an EM fashion. Moreover, to pursue spatiotemporally compact clusters, for each pixel $i\!\in\!\mathcal{I}$, we supply its visual embedding $\bm{i}$ with a 3D sinu- soidal position$_{\!}$ encoding$_{\!}$ vector$_{\!}$~\cite{vaswani2017attention,dosovitskiy2020image}. In practice, only$_{\!}$ a$_{\!}$ small$_{\!}$ number$_{\!}$ of$_{\!}$ EM$_{\!}$ steps$_{\!}$ (\ie,$_{\!}$ 100)$_{\!}$ can$_{\!}$ deliver$_{\!}$ satisfactory clustering results, taking about $2$ seconds per video, averaged on our training dataset -- YouTube-VOS$_{\!}$~\cite{xu2018youtube}.

\noindent\textbf{Step$_{\!}$ 2: Mask-embedded Segmentation Learning.} In this\\
\noindent step,$_{\!}$ our$_{\!}$ model$_{\!}$~utilizes$_{\!\!}$ clustering$_{\!}$ results$_{\!}$ as$_{\!}$ pseudo$_{\!}$ ground- truths$_{\!}$ (see$_{\!}$~Fig.$_{\!}$~\ref{fig:framework}(b)),$_{\!}$ to$_{\!}$ directly$_{\!}$ learn$_{\!}$ VOS$_{\!}$ as$_{\!}$ mask$_{\!}$~embed-  ding and decoding. For each training video $\mathcal{I}$, we sample $N_{\!}+_{\!}1$$_{\!}$ frames$_{\!}$ $\{I_{r_1}, I_{r_2},_{\!} \cdots_{\!}, I_{r_N}, I_{q}\}$ and their masks $\{Y_{r_1}, $ $Y_{r_2}, \cdots, Y_{r_N}, Y_{q}\}$, as$_{\!}$ training$_{\!}$ examples.$_{\!}$ The$_{\!}$ pseudo$_{\!}$ masks are naturally derived from the assignment matrix $\bm{S}^*$, cor- responding to the pixel-level assignment of a certain cluster.$_{\!}$ The$_{\!}$ training$_{\!}$ examples$_{\!}$ are$_{\!}$ used$_{\!}$ to$_{\!}$ teach$_{\!}$~our$_{\!}$ model$_{\!}$ to refer$_{\!}$  to$_{\!}$ the$_{\!}$ first$_{\!}$ $N_{\!}$ frame-mask$_{\!}$ pairs$_{\!\!}$ $\{(I_{r_n\!},_{\!} Y_{r_n\!})\}_{n\!}$  to$_{\!}$ segment$_{\!}$ the$_{\!}$ last$_{\!}$ query$_{\!}$ frame$_{\!}$ $I_{q\!}$ ---$_{\!}$ predicting$_{\!}$ $Y_{q}$.$_{\!}$ As$_{\!}$ such,$_{\!}$ our$_{\!}$ model$_{\!}$~can$_{\!}$ learn$_{\!}$~i)$_{\!}$~\textit{mask embedding}:$_{\!}$ how$_{\!}$ to$_{\!}$ extract$_{\!}$ target-specific$_{\!}$ con- text$_{\!}$ from$_{\!}$ $\{(I_{r_{n\!}},_{\!} Y_{r_n\!})\}_{n}$;$_{\!}$ and$_{\!}$ ii)$_{\!}$~\textit{mask decoding}: how to make use of target-specific context to segment the target in $I_{q}$. 

More specifically, we first respectively apply our visual encoder $\mathcal{E}$ and frame-mask  encoder $\mathcal{V}$ over each reference frame $I_{r_n\!}$ and each reference frame-mask~pair $(I_{r_n}, Y_{r_n})$, to  obtain visual and target-specific embeddings:
\begin{equation}\small\label{eq:memoryencoding}
   \!\!\bm{I}_{r_n\!}\!=\!\mathcal{E}(I_{r_n})\!\in\!\mathbb{R}^{HW\times D},~~~\bm{V}_{r_n\!}\!=\!\mathcal{V}([I_{r_n}, Y_{r_n}])\!\in\!\mathbb{R}^{HW \times D'}.
\vspace{-2pt}
\end{equation}
We respectively stack all the reference visual and target-specific embeddings:~$\bm{I}_{r}\!=$ $[\bm{I}_{r_1},_{\!} \cdots_{\!}, \bm{I}_{r_N}]\!\in\!\mathbb{R}^{NHW\!\times\!D}$, and $\bm{V}_{r}\!=\![\bm{V}_{r_1},_{\!} \cdots_{\!}, \bm{V}_{r_N}]\!\in\!\mathbb{R}^{NHW\!\times\!D'}$. To leverage $\bm{V}_{r\!}$ to boost$_{\!}$ the$_{\!}$ prediction$_{\!}$ of$_{\!}$ $I_{q}$,$_{\!}$ we$_{\!}$ need$_{\!}$ to$_{\!}$ mine$_{\!}$ useful$_{\!}$ context,$_{\!}$ related$_{\!}$ to$_{\!}$ $I_{q}$,$_{\!}$ from$_{\!}$ $\bm{V}_{r}$. Given the visual embedding $\bm{I}_{q}\!\in\!\mathbb{R}^{HW\!\times\!D\!\!}$ of $I_{q\!}$ (extracted from $\mathcal{E}$), we$_{\!}$~estimate$_{\!}$~the affinity between the query $I_{q\!}$ and reference frames $\{I_{r_n}\}_{n\!}$ (analogous to Eq.$_{\!}$~\ref{eq:correspondence}):
\vspace{-4pt}
\begin{equation}\small\label{eq:affinity}
   {A} = \texttt{softmax} (\bm{I}_r\bm{I}_q^\top) ~\in\mathbb{R}^{NHW \times HW}.
\vspace{-3pt}
\end{equation}
Hence target-specific, supportive features are accordingly assembled to yield:
\vspace{-10pt}
\begin{equation}\small\label{eq:read}
   ~~~~~~~~~~~~~~~\bm{V}_q  = {A}^\top\bm{V}_r  ~\in\mathbb{R}^{HW \times D'}.
\vspace{-4pt}
\end{equation}
Here$_{\!}$ $\bm{V}_{q\!}$ absorbs$_{\!}$ existent$_{\!}$ object$_{\!}$ observations$_{\!}$ in$_{\!}$ the$_{\!}$ reference set $\{(I_{r_n\!},_{\!} Y_{r_n\!})\}_{n}$, revealing~for~$I_{q\!}$ whether each pixel thereof  belongs to the target object or not. Given precise~segmentation groundtruths, it is relatively easy for fully supervised methods$_{\!}$~\cite{oh2019video,cheng2021rethinking,seong2020kernelized} to learn to~directly decode $\bm{V}_{q\!}$  into segmentation mask. However, this strategy~does not work well in our case since the pseudo labels are inevitably noisy and\\
\noindent less$_{\!}$ accurate,$_{\!}$ compared$_{\!}$ with$_{\!}$ the$_{\!}$ real$_{\!}$ groundtruths.$_{\!}$ To$_{\!}$ tackle$_{\!}$ this, we achieve mask decoding through  a \textit{mask refinement} scheme, which makes more explicit use of reference masks. Specifically,  we first construct a coarse mask $\bar{Y}_{q\!}$ for $I_{q\!}$ by warping the reference masks $\{Y_{r_n}\}_{n\!}$ \textit{w.r.t.} the affinity~$A$:
\vspace{-4pt}
 \begin{equation}\small\label{eq:initialmask}
   \bar{Y}_q  = {A}^{\top} [Y_{r_1}, Y_{r_2}, \cdots\!, Y_{r_n}] ~\in\mathbb{R}^{HW}.
 \vspace{-3pt}
 \end{equation}
\noindent The segmentation prediction $\hat{Y}_q$ for the query $I_q$ is made as:
\vspace{-5pt}
\begin{equation}\small\label{eq:refinement}
   \hat{Y}_q = \mathcal{D}([\bm{V}_q, \bar{\bm{V}}_q]), ~~~~~~\bar{\bm{V}}_q = \mathcal{V}([I_q, \bar{Y}_q])~\in\mathbb{R}^{HW \times D'}.
   \vspace{1pt}
\end{equation}
Here$_{\!}$ the$_{\!}$ frame-mask$_{\!}$ encoder$_{\!}$ $\mathcal{V}_{\!}$ (\textit{cf.}$_{\!}$~Eq.$_{\!}$~\ref{eq:memoryencoding}) is smartly revoked to get another target-specific embedding $\bar{\bm{V}}_q$, from the pair of the query frame $I_q$ and warped coarse mask $\bar{Y}_q$. This also elegantly resembles the mask copying strategy adopted in existing$_{\!}$ correspondence-based$_{\!}$ self-supervised$_{\!}$ VOS$_{\!}$ models.$_{\!}$ Conditioned on the concatenation of $\bm{V}_q$  and $\bar{\bm{V}}_q$, the mask decoder $\mathcal{D}$ outputs a finer mask $ \hat{Y}_q$. In practice we find our mask refinement strategy can ease training and bring better

\noindent performance (related experiments can be found in Table$_{\!}$~\ref{table:ablation5}).

Given the pseudo segmentation label $Y_q$ and prediction $\hat{Y}_q$ of $I_q$, our whole model is supervised by minimizing the standard cross-entropy loss $\mathcal{L}_{\text{CE}}$:
\vspace{-2pt}
\begin{equation}\small\label{eq:celoss}
	\mathcal{L}_{\text{Seg}} = \sum\nolimits_{\mathcal{I}}\mathcal{L}_{\text{CE}} ( \hat{Y}_q, Y_q).
\vspace{-1pt}
\end{equation}

\subsection{Self-supervised$_{\!}$ Dense$_{\!}$ Correspondence$_{\!}$ Learning$_{\!\!\!\!\!\!\!}$}\label{sec:correspondence}\vspace{-1pt}
An appealing aspect of  our mask embedding framework is that it is general enough to naturally incorporate unsupervised correspondence learning to specifically reinforce visual representation $\mathcal{E}$. This comes with a few advantages: First, this allows our model to exploit the inherent coherence in natural videos as free supervisory signals to promote the transferability and sharpen the discriminativeness of  $\mathcal{E}$. Second, correspondence learning provides initial meaningful features for clustering$_{\!}$ (\textit{cf.}$_{\!}$~Eq.$_{\!}$~\ref{eq:kmeans}),$_{\!}$ which is prone to degeneracy (\ie, allocating most samples to the  same cluster) caused by poor initialization$_{\!}$~\cite{caron2018deep}. Third, our segmentation model involves the computation of intra-frame affinity $A$ ($_{\!}$\textit{cf.}$_{\!}$~Eqs.$_{\!}$~\ref{eq:memoryencoding}-\ref{eq:read}), raising a strong demand for efficiently modeling dense correspondence within our framework. Along with recent work of {contrastive matching} based correspondence learning \cite{jeon2021mining,araslanov2021dense,xu2021rethinking}, we comprehensively explore intrin- sic$_{\!}$ continuity$_{\!}$ within$_{\!}$ raw$_{\!}$ videos$_{\!}$ in$_{\!}$ both$_{\!}$ \textit{short-term}$_{\!}$ and$_{\!}$ \textit{long-term} time scales, to boost the learning of $\mathcal{E}_{\!}$ (see$_{\!}$~Fig.$_{\!}$~\ref{fig:framework}(c)).

\noindent\textbf{Short-term$_{\!}$ Appearance$_{\!}$ Consistency.}$_{\!}$ Temporally$_{\!}$ adjacent frames typically exhibit continuous and trivial appearance changes \cite{hurri2003simple,lu2020learning}. To accommodate this~property, we~enforce
\textit{transformation-equivariance}$_{\!}$~\cite{novotny2018self,thewlis2017unsupervised,thewlis2017unsupervised2,o2020unsupervised} between our adjacent frame representations. Given two \textbf{successive} frames ${I}_{t}, {I}_{t+1}\!\in\!\mathcal{I}$, their representations, delivered by $\mathcal{E}$, are cons- trained to
be \textbf{equivariant} against geometric transformations\\
\noindent (\ie, scaling, flipping, and cropping). Specifically, denote $\Phi$ as~a random transformation, our \textit{equivariance based short- term$_{\!}$ appearance$_{\!}$ consistency$_{\!}$ constraint}$_{\!}$ can$_{\!}$ be$_{\!}$ expressed$_{\!}$ as:
\vspace{-4pt}
\begin{equation}
\small\label{eq:spatial}
\begin{aligned}
\!\!\!\!\!\!\!\text{\hypertarget{Q1}{\ding{182}}}&~\mathcal{E}(I_{t}) \approx \mathcal{E}(I_{t+1})\\[-5pt]
&\text{\footnotesize\!\!\!\!\!\!{\color{gray}short-term consistency}}\\[1.5pt]
\!\!\!\!\!\!\!\text{\hypertarget{Q2}{\ding{183}}}&~\mathcal{E}(\Phi(I_{t})) = \Phi(\mathcal{E}(I_{t}))\\[-5pt]
&\text{\footnotesize\!\!\!\!\!\!{\color{gray}transformation-equivariance}}
\end{aligned}
\left.
\begin{aligned}
\\
\\
\end{aligned}
\right\}\!\Rightarrow\!\mathcal{E}(\Phi(I_{t})) \approx \Phi(\mathcal{E}(I_{t+1}))~~\text{\hypertarget{Q3}{\ding{184}}}.\!\!
\vspace{-1pt}
\end{equation}
Here \myhyperlink{Q1}{\ding{182}} states the short-term consistency property; \myhyperlink{Q2}{\ding{183}} refers to the equivariance~constraint on a single image$_{\!}$~\cite{o2020unsupervised}, \ie, an imagery transformation $\Phi$ of $I_{t}$ should lead to a correspondingly transformed feature$_{\!}$~\cite{araslanov2021dense}. By bringing \myhyperlink{Q2}{\ding{183}} into \myhyperlink{Q1}{\ding{182}}, we prevent trivial solution, \ie, $\mathcal{E}(I_{t})\!\equiv\!\mathcal{E}(I_{t+1})$, when directly optimizing $\mathcal{E}$ via \myhyperlink{Q1}{\ding{182}}, and eventually get \myhyperlink{Q3}{\ding{184}}.

Following$_{\!}$ \myhyperlink{Q3}{\ding{184}},$_{\!}$ we$_{\!}$ first$_{\!}$ get$_{\!}$ the$_{\!}$ feature$_{\!}$ of$_{\!}$ transformed$_{\!}$ ${I}_{t}$:$_{\!\!}$ $\bm{X}'_{t\!}\!=\!\mathcal{E}(\Phi(I_{t}))_{\!}\!\in_{\!}\!\mathbb{R}^{HW\!\times\!D\!}$,$_{\!}$ and$_{\!}$ transformed$_{\!}$ feature$_{\!}$ of$_{\!}$ ${I}_{t+1\!}$:$_{\!\!}$ $\bm{X}_{t+1\!}\!=_{\!}\!\Phi(\mathcal{E}(I_{t+1}))_{\!}\!\in_{\!}\!\mathbb{R}^{HW\!\times\!D}$.$_{\!}$ Denote$_{\!}$ $k$-\textit{th}$_{\!}$ pixel$_{\!}$ feature of $\bm{X}_{t+1\!}$ (resp.$_{\!}$ $\bm{X}'_{t}$) as $\bm{x}^k_{t+1}\!\in\!\mathbb{R}^{D}$ (resp.$_{\!}$ $\bm{x}'^k_{t}\!\in\!\mathbb{R}^{D}$)\footnote{For clarity, the symbols for frame and pixel features in \S\ref{sec:correspondence} are slightly redefined as $\bm{X}$ and $\bm{x}$, instead of using $\bm{I}$ and $\bm{i}$ as in \S\ref{sec:framework}.}, our short-term consistency loss is computed as:
\vspace{-3pt}
\begin{equation}\small\label{eq:consistency}
\begin{aligned}
	\mathcal{L}_\text{Short} = -\sum\nolimits_{\mathcal{I}}\sum\nolimits_{k} \log \frac{\exp(\langle\bm{x}^{k\top}_{t+1}\bm{x}'^{k}_{t}\rangle)}{\sum_{l}\exp(\langle\bm{x}^{k\top}_{t+1}\bm{x}'^{l}_{t}\rangle)},
\end{aligned}
\vspace{-2pt}
\end{equation}
where $\langle\bm{x}^{k\top}_{t+1}\bm{x}'^{l}_{t}\rangle$ gives cosine similarity based affinity between $k$-\textit{th} pixel feature~of $\bm{X}_{t+1\!}$ and $l$-\textit{th} pixel feature of $\bm{X}'_{t}$.$_{\!}$ Eq.$_{\!}$~\ref{eq:consistency}$_{\!}$ captures$_{\!}$ local$_{\!}$ appearance$_{\!}$ continuity$_{\!}$ by$_{\!}$ contras- ting$_{\!}$ affinity$_{\!}$ between$_{\!}$ aligned$_{\!}$~pixel$_{\!}$ feature$_{\!}$ pairs,$_{\!}$ \ie,$_{\!}$ $\bm{x}^{k}_{t+1\!\!}$ and$_{\!}$ $\bm{x}'^{k\!}_{t}$  against$_{\!}$ non-corresponding$_{\!}$ ones,$_{\!}$ \ie,$_{\!}$ $\bm{x}^{k}_{t+1\!}$ and$_{\!}$ $\{\bm{x}'^{l}_{t}\}_{l\neq k}$, with an extra transformation equivariance based constraint. 

\noindent\textbf{Long-term Semantic Dependency.} In addition to considering the local consistency among adjacent frames, we exploit long-term coherence of visual content among distant frames \cite{mobahi2009deep,yan2019fine}. To address this property, we enforce trans-\\
\noindent formation$_{\!}$ equivariance$_{\!}$ between$_{\!}$ representations$_{\!}$ of$_{\!}$ \textit{arbitrary}$_{\!}$ frame$_{\!}$ pairs$_{\!}$ (sampled$_{\!}$ from$_{\!}$ the$_{\!}$ same$_{\!}$ video$_{\!}$) after$_{\!}$ \textit{alignment}. Given$_{\!}$ two$_{\!}$ \textbf{distant}$_{\!}$ frames$_{\!}$ ${I}_{t},_{\!} {I}_{t'}\!\in\!\mathcal{I}$ (\textit{s.t.}$_{\!}$~$|t_{\!}-_{\!}t'|_{\!}\geq_{\!}5$),$_{\!}$ their representations, after being \textbf{aligned} \textit{w.r.t.} their affinity~$A_{t'}^{t}$, are~constrained to be \textbf{equivariant} against geometric transformations. In particular, denote $A_{t'\!}^{t}\in_{\!}[0,1]^{HW\!\times\!HW}$ (resp.\\
\noindent $A_{\Phi(t')\!\!}^{t}\in_{\!\!}[0,1]^{HW\!\times\!HW}$)$_{\!}$ as$_{\!}$ the$_{\!}$ affinity$_{\!}$ between$_{\!}$ $I_{t\!}$ and$_{\!}$ $I_{t'\!}$~(resp.$_{\!\!}$ $I_{t}$ and $\Phi(I_{t'})$),  our \textit{equivariance based long-term semantic dependency constraint} can be expressed as:
\vspace{-2pt}
\begin{equation}
\small\label{eq:temporal}
\begin{aligned}
\!\!\!\!\!\!\!\!\!\!\!\text{\hypertarget{Q4}{\ding{185}}}&~\mathcal{E}(I_{t})_{\!} \approx_{\!} A_{t'}^{t\top\!}\mathcal{E}(I_{t'})_{\!}\\[-5pt]
&\text{\footnotesize\!\!\!\!\!\!{\color{gray}long-term dependency}}\\[1.5pt]
\!\!\!\!\!\!\!\!\!\!\!\text{\hypertarget{Q2}{\ding{183}}}&~\mathcal{E}(\Phi(I_{t}))_{\!} =_{\!} \Phi(\mathcal{E}(I_{t}))_{\!}\\[-5pt]
&\text{\footnotesize\!\!\!\!\!\!{\color{gray}transformation-equivariance}}
\end{aligned}
\left.
\begin{aligned}
\\
\\
\end{aligned}
\right\}_{\!\!}\Rightarrow_{\!}\mathcal{E}(I_{t})_{\!} \approx_{\!}  A_{\Phi(t')}^{t\top\!}\Phi(\mathcal{E}(I_{t'}))~\text{\hypertarget{Q5}{\ding{186}}}.\!\!
\vspace{-2pt}
\end{equation}
Here \myhyperlink{Q4}{\ding{185}} states the long-term dependency property; \myhyperlink{Q2}{\ding{183}} poses the$_{\!}$ equivariance$_{\!}$ constraint,$_{\!}$ as$_{\!}$ in$_{\!}$ Eq.$_{\!}$~\ref{eq:spatial}.$_{\!}$ By$_{\!}$ bringing$_{\!}$~\myhyperlink{Q2}{\ding{183}}$_{\!}$ into$_{\!}$ \myhyperlink{Q4}{\ding{185}},$_{\!}$~we$_{\!}$ prevent$_{\!}$ trivial$_{\!}$ solution,$_{\!}$ \ie,$_{\!}$ $\mathcal{E}(I_{t})\!\equiv\!\mathcal{E}(I_{t'})$,$_{\!}$ when di- rectly$_{\!}$ optimizing$_{\!}$ $\mathcal{E}_{\!}$ via$_{\!}$ \myhyperlink{Q4}{\ding{185}},$_{\!}$ and$_{\!}$ eventually$_{\!}$ get$_{\!}$ \myhyperlink{Q5}{\ding{186}}.$_{\!}$ Specifically,$_{\!}$ similar to \myhyperlink{Q4}{\ding{185}}, we have $\mathcal{E}(I_{t})_{\!} \approx_{\!} A_{\Phi(t')}^{t\top\!}\mathcal{E}(\Phi(I_{t'}))$; then with \myhyperlink{Q2}{\ding{183}}, we
obtain $\mathcal{E}(I_{t})_{\!} \approx_{\!} A_{\Phi(t')}^{t\top\!}\mathcal{E}(\Phi(I_{t'}))_{\!}=_{\!}A_{\Phi(t')}^{t\top\!}\Phi(\mathcal{E}(I_{t'}))$.

$_{\!\!}$Following$_{\!}$ \myhyperlink{Q5}{\ding{186}},$_{\!}$ we$_{\!}$ get$_{\!}$ the$_{\!}$ feature$_{\!}$ of$_{\!}$ transformed$_{\!}$ ${I}_{t'}$:$_{\!}$ $\bm{X}'_{t'\!}\!=\!\mathcal{E}(\Phi(I_{t'}))\!\in\!\mathbb{R}^{HW\!\times\!D}$,$_{\!}$ transformed feature of ${I}_{t'}$: $\bm{X}_{t'}\!=\!\Phi(\mathcal{E}(I_{t'}))\!\in\!\mathbb{R}^{HW\!\times\!D}$, and the original feature of ${I}_{t}$: $\bm{I}_{t}\!=\!\mathcal{E}(I_{t})\!\in\!\mathbb{R}^{HW\!\times\!D}$. For $k$-\textit{th} pixel (feature) of $\bm{X}'_{t'}$, we first find$_{\!}$~the matching (\ie, the most similar) pixel $o_{k\!}$ in $\bm{I}_{t}$~as:
\vspace{-2pt}
\begin{equation}\small\label{eq:assignment}
	o_k = \underset{o \in \{1, \cdots_{\!}, HW\}}{\arg \max } a_{k,o}, ~~~~~~ a_{k,o} = \frac{\exp(\langle\bm{x}'^{k\top}_{t'}\bm{i}^{o}_{t}\rangle)}{\sum\nolimits_{l} \exp(\langle\bm{x}'^{k\top}_{t'}\bm{i}^{l}_{t}\rangle)},
\vspace{-3pt}
\end{equation}
where $\bm{i}^{o}_{t}\!\in\!\mathbb{R}^{D\!}$ refers to $o$-\textit{th} pixel feature of $\bm{I}_{t}$, and $a_{k,o}$ corresponds to $(k,o)$-\textit{th} element of the affinity $A_{\Phi(t')\!}^{t}$ between $\Phi(I_{t'})$ and $I_{t}$. Then, the dominant index ok serves
as pseudo labels for our temporally-distant matching and our long-term dependency loss is computed as:  \vspace{-2pt}
\begin{equation}\small\label{eq:dependency}
	\mathcal{L}_\text{Long} = -\sum\nolimits_{\mathcal{I}}\sum\nolimits_{k} \log \frac{\exp(\langle\bm{x}^{k\top}_{t'}\bm{i}^{o_k}_{t}\rangle)}{\sum_{l}\exp(\langle\bm{x}^{k\top}_{t'}\bm{i}^{l}_{t}\rangle)}.
\vspace{-1pt}
\end{equation}
Eq.$_{\!}$~\ref{eq:dependency} addresses global semantic dependencies by contrasting affinity between aligned pixel feature pairs, \ie, $\bm{x}^{k}_{t'\!}$ and $\bm{i}^{o_k}_{t}$, against non-corresponding ones, \ie, $\bm{x}^{k}_{t'\!}$ and $\{\bm{i}^{l}_{t}\}_{l\neq o_k}$, under an equivariant representation learning scheme.

\begin{figure*}
\begin{minipage}{\textwidth}
\begin{minipage}[t]{0.30\textwidth}
\subsection{Implementation Details}\label{sec:Id}
  \vspace{-2pt}

\noindent\textbf{Full$_{\!}$ Loss.$_{\!}$} The$_{\!}$ overall$_{\!}$ training$_{\!}$ loss$_{\!}$ is:
\vspace{-8pt}
\begin{equation}\small\label{eq:loss}
\begin{aligned}
   \mathcal{L}=& \mathcal{L}_{\text{Seg}}+\mathcal{L}_{\text{Corr}}\\=&\mathcal{L}_{\text{Seg}}+\lambda_1\mathcal{L}_{\text{Short}} +\lambda_2\mathcal{L}_{\text{Long}},
   \vspace{-1pt}
\end{aligned}
\end{equation}
where$_{\!}$ the$_{\!}$ coefficients$_{\!}$ are$_{\!}$ empirically set as: $\lambda_{1\!}=_{\!}0.1$ and $\lambda_{2\!}=_{\!}0.5$.

\noindent\textbf{Network$_{\!}$ Configuration.$_{\!}$} For$_{\!}$ the$_{\!}$ \textit{vi- sual$_{\!}$ encoder}$_{\!}$ $\mathcal{E}$,$_{\!}$ we$_{\!}$ instantiate$_{\!}$ it$_{\!}$ as$_{\!}$ \texttt{ResNet-18}  or \texttt{ResNet-50} in our experiments.  For \texttt{ResNet-18}, the spatial strides of the second and last residual blocks are removed to yield an$_{\!}$ output$_{\!}$ stride$_{\!}$ of$_{\!}$ $8$,$_{\!}$ as$_{\!}$ in$_{\!}$~\cite{jabri2020space,wang2021contrastive,araslanov2021dense}.$_{\!\!}$
For \texttt{ResNet-50}, we follow \cite{wang2019learning} to take features from \texttt{res4}, and drop its stride to preserve more spatial details.$_{\!}$ For$_{\!}$ the$_{\!}$ \textit{frame-mask$_{\!}$ encoder}$_{\!}$ $\mathcal{V}$,$_{\!}$ it$_{\!}$ has a~similar structure as $\mathcal{E}$, expect for the input$_{\!}$ and$_{\!}$ output$_{\!}$ dimensionality.$_{\!}$ On$_{\!}$ the$_{\!}$ top$_{\!}$ of $\mathcal{E}$ and $\mathcal{V}$, two $1\!\times\!1$ convolution layers are separately added
   \end{minipage}
   \begin{minipage}[t]{0.005\textwidth}
   ~~~~~~
   \end{minipage}
    \begin{minipage}[t]{0.68\textwidth}
    \vspace{-2pt}
    \begin{threeparttable}
        \resizebox{0.99\textwidth}{!}{
		\setlength\tabcolsep{2pt}
      \renewcommand\arraystretch{1.03}
      \begin{tabular}{z{68}y{25}crlccccc}
         \toprule[1.0pt]
         \multicolumn{2}{c}{Method} & Backbone  & Dataset &\!\!(size) & $\!\!\mathcal{J}$\&$\mathcal{F}_m$$_{\!\!\!}$ $\uparrow$ & $\mathcal{J}_m$$_{\!\!\!}$ $\uparrow$ & $\mathcal{J}_r$$_{\!\!\!}$ $\uparrow$  &  $\mathcal{F}_m$$_{\!\!\!}$ $\uparrow$  & $\mathcal{F}_r$$_{\!\!\!}$ $\uparrow$  \\ \hline
         Colorization\!~\cite{vondrick2018tracking}&\!\!\pub{ECCV18} & \texttt{ResNet-18}  & Kinetics &\!\!(~-~, 800 hours)
         & 34.0 & 34.6 & 34.1 & 32.7 & 26.8 \\
         CorrFlow\!~\cite{lai2019self}&\!\!\pub{BMVC19} & \texttt{ResNet-18}  & OxUvA &\!\!(~-~, 14 hours)
         & 50.3 & 48.4 & 53.2 & 52.2 & 56.0 \\
         TimeCycle\!~\cite{wang2019learning}&\!\!\pub{CVPR19} & \texttt{ResNet-50}  & VLOG &\!\!(~-~, 344 hours)
         & 48.7 & 46.4 & 50.0 & 50.0 & 48.0 \\
         UVC\!~\cite{li2019joint}&\!\!\pub{NeurIPS19} & \texttt{ResNet-18}  & C+Kinetics &\!\!(30K, 800 hours)
         & 57.8 & 56.3 & 65.0 & 59.2 & 64.1 \\
         MuG\!~\cite{lu2020learning}&\!\!\pub{CVPR20} & \texttt{ResNet-18} & OxUvA &\!\!(~-~, 14 hours)
         & 54.3 & 52.6 & 57.4 & 56.1 & 58.1  \\
         MAST\!~\cite{lai2020mast}&\!\!\pub{CVPR20} & \texttt{ResNet-18}  & Youtube-VOS &\!\!(~-~, 5.58 hours)
         & 65.5 & 63.3 & 73.2 & 67.6 & 77.7  \\
         CRW\!~\cite{jabri2020space}&\!\!\pub{NeurIPS20} & \texttt{ResNet-18}  & Kinetics &\!\!(~-~, 800 hours)
         & 68.3 & 65.5 & 78.6 & 71.0 & 82.9 \\
         ConCorr\!~\cite{wang2021contrastive}&\!\!\pub{AAAI21} & \texttt{ResNet-18}  &C+TrackingNet &\!\!(30K, 300 hours)\!\!\!\!
         & 63.0 & 60.5 & 70.6 & 65.5 & 73.0 \\
         CLTC\!~\cite{jeon2021mining}&\!\!\pub{CVPR21}  & \texttt{ResNet-18} & Youtube-VOS &\!\!(~-~, 5.58 hours)
         & 70.3 & 67.9 & 78.2 & 72.6 & 83.7 \\
         JSTG\!~\cite{zhao2021modelling}&\!\!\pub{ICCV21} & \texttt{ResNet-18} & Kinetics &\!\!(~-~, 800 hours)
         & 68.7 & 65.8 & 77.7  & 71.6 & 84.3 \\
         & & \texttt{ResNet-18} &  &
         & 67.9 & 65.0 & 77.2 & 70.8 & 82.3 \\
         \multirow{-2}{*}{VFS \!~\cite{xu2021rethinking}}&\multirow{-2}{*}{\!\!\pub{ICCV21}} & \texttt{ResNet-50} & \multirow{-2}{*}{Kinetics} &\!\!\multirow{-2}{*}{(~-~, 800 hours)}
         & 69.4 & 66.7 & 78.6 & 72.0 & 85.2 \\
         & & \texttt{ResNet-50} &  &&56.2 &54.5 &58.1 &57.9 &60.3 \\
         \multirow{-2}{*}{DINO\!~\cite{caron2021emerging}}&\multirow{-2}{*}{\!\!\pub{ICCV21}} & \texttt{ViT-B/8} &\multirow{-2}{*}{I} &\!\!\multirow{-2}{*}{(1.28M, ~-~)} & 71.4 & 67.9 & 81.6 & 74.9 & 85.4 \\
         DUL\!~\cite{araslanov2021dense}&\!\!\pub{NeurIPS21} & \texttt{ResNet-18}  & Youtube-VOS &\!\!(~-~, 5.58 hours)
         & 69.3 & 67.1 & 81.2 & 71.6 & 84.9 \\
         SCR\!~\cite{son2022contrastive}&\!\!\pub{CVPR22} & \texttt{ResNet-18}  & Kinetics &\!\!(~-~, 800 hours)
         & 70.5 & 67.4 & 78.8 & 73.6 & 84.6 \\
         LIIR\!~\cite{li2022locality}&\!\!\pub{CVPR22} & \texttt{ResNet-18}  & Youtube-VOS &\!\!(~-~, 5.58 hours)
         & 72.1 & 69.7 & 81.4 & 74.5 & 85.9 \\
         \hline
         && \texttt{ResNet-18} &
         && \textbf{74.5} & \textbf{71.6} & \textbf{82.9} &  \textbf{77.4} & \textbf{86.9} \\
         \multirow{-2}{*}{\textbf{\textsc{Ours}}} && \texttt{ResNet-50}  &\multirow{-2}{*}{Youtube-VOS} &\!\!\multirow{-2}{*}{(~-~, 5.58 hours)}
         & \textbf{75.6} & \textbf{73.3} & \textbf{83.6} &  \textbf{77.8} & \textbf{87.3} \\
         \hline
          \!\color{mygray2}OSVOS\!~\cite{caelles2017one}&\!\!\pub{CVPR17}  & \texttt{\color{mygray2}VGG-16} & \color{mygray2}I+D &\!\!\color{mygray2}(1.28M, 10k) & \color{mygray2}60.3 & \color{mygray2}56.6 & \color{mygray2}63.8 & \color{mygray2}63.9 & \color{mygray2}73.8 \\

         \!\color{mygray2}STM\!~\cite{oh2019video}&\!\!\pub{ICCV19} & \texttt{\color{mygray2}ResNet-50} & \color{mygray2}I+D+Youtube-VOS &\!\!\color{mygray2}(1.28M, 164k)
          & \color{mygray2}81.8 & \color{mygray2}79.2 & \color{mygray2}88.7 & \color{mygray2}84.3 & \color{mygray2}91.8 \\

          \bottomrule[1.0pt]
      \end{tabular}
        }
        \leftline{{\footnotesize{~~~~~~~- I:~ImageNet$_{\!}$~\cite{deng2009imagenet}; C:~COCO$_{\!}$~\cite{lin2014microsoft}; D:~DAVIS$_{17\!}$~\cite{perazzi2016benchmark}.}}}
    \end{threeparttable}
    \vspace*{-7pt}
    \captionsetup{width=.92\textwidth}
    \makeatletter\def\@captype{table}\makeatother\captionsetup{font=small}\caption{\small\textbf{Quantitative segmentation results} (\S\ref{sec:performance}) on DAVIS$_{17}$\!~\cite{perazzi2016benchmark} \texttt{val}. For dataset size, we report (\#\textit{raw} images, length of \textit{raw} videos) for self-supervised methods and (\#image-level annotations, \#pixel-level annotations) for {\color{mygray2}supervised} methods.
    \label{results:davis}}
  \end{minipage}
  \end{minipage}
  \vspace*{-21pt}
\end{figure*}

\noindent   to reduce~the~output dimensions of $\mathcal{E}$ and $\mathcal{V}$  to  $D\!=\!128$ and $D'\!=\!512$, respectively. For the  \textit{mask decoder} $\mathcal{D}$, it consists  of two Residual blocks that are connected with $\mathcal{E}$ through skip layers, and a $1\!\times\!1$ convolution layer to produce the final  segmentation  prediction.

\noindent\textbf{Training.}$_{\!}$ We follow~\cite{caron2021emerging} to pre-train the backbone network $\mathcal{E}$ on YouTube-VOS for 300 epochs, enabling reliable$_{\!}$ initial clustering. Then, we$_{\!}$ conduct the main training for$_{\!}$ a$_{\!}$ total$_{\!}$ of$_{\!}$ $400_{\!}$ epochs$_{\!}$ using Adam optimizer with batch size $16$ and base learning rate 1e-4, on one Tesla A100 GPU. In the first $300$ epochs, the$_{\!}$ whole$_{\!}$ network$_{\!}$ is$_{\!}$ trained$_{\!}$ with$_{\!}$ only$_{\!}$ the$_{\!}$ correspondence$_{\!}$~loss $\mathcal{L}_{\text{Corr}}$.$_{\!}$
 The$_{\!}$ learning$_{\!}$ rate$_{\!}$ is$_{\!}$ scheduled$_{\!}$ following$_{\!}$~a$_{\!}$ ``step''$_{\!}$ policy, decayed by$_{\!}$ multi-plying$_{\!}$ $0.5$ every $100$ epochs. In the last $100$ epochs, the whole network~is  trained$_{\!}$ using$_{\!}$ the$_{\!}$ full$_{\!}$ loss$_{\!}$ $\mathcal{L}$,$_{\!}$ with$_{\!}$ fixed$_{\!}$ learning$_{\!}$ rate$_{\!}$ 1e-5. The first time-space clustering is made at epoch $300$ for creating initial pseudo segmentation labels. Afterwards, the  pseudo labels are updated by conducting re-clustering$_{\!}$ at$_{\!}$ every$_{\!}$ $10$$_{\!}$ epochs. During clustering, we abandon$_{\!}$ over-size$_{\!}$ clusters,$_{\!}$ \ie, accounting$_{\!}$ for$_{\!}$ more$_{\!}$ than$_{\!}$ 40\%$_{\!}$ of$_{\!}$ video$_{\!}$ pixels.$_{\!}$ These$_{\!}$ big$_{\!}$ clusters$_{\!}$ are$_{\!}$ typically$_{\!}$ scene$_{\!}$ background, like~sky and grass; only the remaining pixel clusters/segments are used as pseudo labels. Random scaling, cropping, and flipping  are used for data augmentation, and the training image size is set to  $256\!\times\!256$.
In each mini-batch, we sample $3$ frames per video, and adopt the strategy in \cite{cheng2021rethinking,oh2019video} to learn  mask decoding with two reference frames (\ie, $N\!=\!2$).

\vspace{-1pt}
\noindent\textbf{Testing.} Once trained, our model is applied to test videos without any fine-tuning. Following~\cite{jabri2020space,araslanov2021dense}, for each query frame, we take the first frame (providing reliable object mask information), and, if applicable, its prior 20 frames (capturing diverse object patterns), as well as their masks, as reference for segmentation prediction. In addition, we repeatedly feed the prediction $\hat{Y}_q$  back to the mask decoder $\mathcal{D}$ for iterative refinement. We find this strategy brings better results while requiring no extra parameters, with only marginal sacrifice of inference speed (see Table$_{\!}$~\ref{table:ablation5}).

  \vspace{-3pt}
\section{Experiments}
  \vspace{-2pt}
\noindent\textbf{Dataset.}$_{\!}$ We$_{\!}$ evaluate$_{\!}$ our$_{\!}$ approach$_{\!}$ on$_{\!}$ two$_{\!}$ VOS$_{\!}$ datasets,$_{\!}$ \ie,

\noindent DAVIS$_{17}$\!~\cite{perazzi2016benchmark}$_{\!}$ and$_{\!}$ YouTube-VOS\!~\cite{xu2018youtube}.$_{\!}$ They$_{\!}$ have$_{\!}$ $30$$_{\!}$ and$_{\!}$ $474$ videos in \texttt{val} sets, respectively. The videos are companied with pixel-wise annotations and cover various challenges like occlusion, complex background, and motion blur.

\noindent\textbf{Evaluation Metric.} Following the official  evaluation  protocols\!~\cite{perazzi2016benchmark,xu2018youtube}, we adopt region similarity ($\mathcal{J}_m$), contour accuracy ($\mathcal{F}_m$) and
their average ($\mathcal{J}$\&$\mathcal{F}_m$). For DAVIS$_{17}$, we additionally report the recall values ($\mathcal{J}_r$ and $\mathcal{F}_r$), at IoU threshold $0.5$.   For YouTube-VOS, scores are obtained by submitting the results to the official evaluation server and separately computed for  \textit{seen} and \textit{unseen} categories.

\begin{figure*}[b]
\vspace{-7pt}
\centering
      \includegraphics[width=0.99\linewidth]{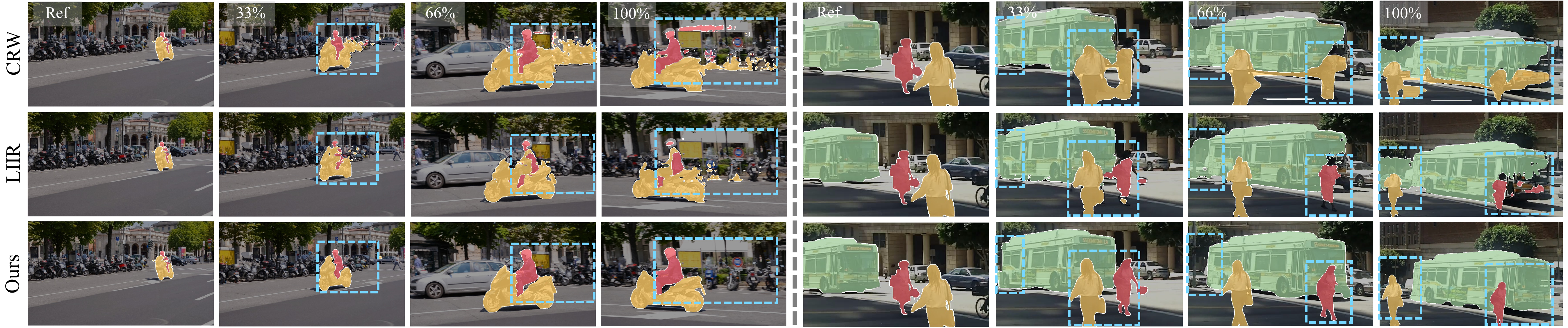}
\vspace{-6pt}
\caption{\textbf{Visual comparison results} (\S\ref{sec:performance}) on
two videos from DAVIS$_{17}$$_{\!}$~\cite{perazzi2016benchmark}$_{\!}$~\texttt{val} (left) and Youtube-VOS$_{\!}$~\cite{xu2018youtube} \texttt{val} (right), respectively.  CRW$_{\!}$~\cite{jabri2020space} and LIIR$_{\!}$~\cite{li2022locality}$_{\!}$ suffer$_{\!}$ from error$_{\!}$ accumulation$_{\!}$ during$_{\!}$ mask$_{\!}$ tracking, due to the simple matching-based mask copy-paste strategy. However, our$_{\!}$ approach$_{\!}$ performs robust over time and yields more accurate segmentation results, by learning to embed target masks.}
\vspace{-5pt}
\label{fig:compa}
\end{figure*}

  \vspace{-3pt}
\subsection{Comparison with State-of-the-Art} \label{sec:performance}
  \vspace{-1pt}
\noindent\textbf{$_{\!}$Performance$_{\!\!}$ on$_{\!\!}$ DAVIS$_{17}$.$_{\!\!}$}
{Table$_{\!\!}$~\ref{results:davis}}$_{\!\!}$~gives$_{\!}$ comparison$_{\!}$ results$_{\!\!}$ against$_{\!}$ 15$_{\!}$~recent self-supervised$_{\!}$ VOS$_{\!}$ methods$_{\!}$ on$_{\!}$ DAVIS$_{17\!}$ \texttt{val}.$_{\!}$ We$_{\!}$ also$_{\!}$ include$_{\!}$ two$_{\!}$ famous$_{\!}$~supervised$_{\!}$  alternatives$_{\!}$~\cite{caelles2017one,oh2019video}$_{\!}$  for$_{\!}$  reference.$_{\!}$  As$_{\!}$  seen,$_{\!}$  using$_{\!}$  a$_{\!}$  relatively$_{\!}$  small$_{\!}$  amount$_{\!}$  of$_{\!}$ training$_{\!}$ data$_{\!}$ (\ie,$_{\!}$ 5.58$_{\!}$ hours$_{\!}$ of$_{\!}$ raw$_{\!}$ videos$_{\!}$ in$_{\!}$ YouTube- VOS~\texttt{train}) and weak backbone architecture  --  \texttt{ResNet- 18}, our approach outperforms all competitors across multiple evaluation$_{\!}$ metrics.$_{\!}$ When$_{\!}$ adopting$_{\!}$ \texttt{ResNet-50},$_{\!}$ our approach yields far better performance, up to \textbf{75.6}\% $\mathcal{J}$\&$\mathcal{F}_m$.

\noindent In$_{\!}$ particular,$_{\!}$ compared with$_{\!}$ \texttt{ResNet-18}$_{\!}$ based$_{\!}$ top-leading$_{\!}$ models,$_{\!}$ \ie,$_{\!}$ LIIR$_{\!}$~\cite{li2022locality}, SCR$_{\!}$~\cite{son2022contrastive}, DUL$_{\!}$~\cite{araslanov2021dense}, and CLTC$_{\!}$~\cite{jeon2021mining},  our approach earns \textbf{2.4}\%, \textbf{4.0}\%, \textbf{5.2}\%, and \textbf{4.2}\% $\mathcal{J}$\&$\mathcal{F}_m$ gains, respectively. Note that,~CLTC~adopts different net-
 work$_{\!}$ architectures$_{\!}$ and$_{\!}$ model$_{\!}$ weights$_{\!}$ for$_{\!}$ different$_{\!}$ datasets.$_{\!}$ Apart$_{\!}$ from$_{\!}$ this, VFS$_{\!}$ and$_{\!}$ JSTG make~use~of much more  training$_{\!}$ data$_{\!}$ than$_{\!}$ ours$_{\!}$ (800~\vs\!\!~5.58$_{\!}$ {hours}$_{\!}$~of videos).$_{\!}$~As$_{\!}$~for DINO,$_{\!}$ a$_{\!}$ recent$_{\!}$ state-of-the-art,$_{\!}$ contrastive$_{\!}$ image$_{\!}$ represen- tation$_{\!}$ learning$_{\!}$ based method, our approach still outperforms it by \textbf{3.1}\% and \textbf{4.2}\% $\mathcal{J}$\&$\mathcal{F}_m$ based on \texttt{ResNet-18} and \texttt{ResNet-50}, respectively.$_{\!}$ This is parti- cularly impressive, considering our backbone is \textit{desperately} \textit{inferior} to DINO (\ie, \texttt{ResNet-18/-50}\!~\vs\!\!~\texttt{ViT-B}) and the training data used by these two methods are completely not comparable in both quality and quantity (\ie, 3.5K videos \vs$_{\!\!}$\!~$1.28$M$_{\!}$ images).$_{\!}$~When$_{\!}$ using$_{\!}$~the$_{\!}$ same$_{\!}$ \texttt{ResNet-50}$_{\!}$ back- bone, the performance gap is huge, \eg,  \textbf{19.4\%} in $\mathcal{J}$\&$\mathcal{F}_m$. Table$_{\!}$~\ref{results:testdev} reports our performance on DAVIS$_{17\!}$ \texttt{test-dev}. We can clearly observe that, our approach, again, suppresses all the recent alternatives by a solid margin.

\begin{table}[t]
   \resizebox{1.01\columnwidth}{!}{
      \setlength\tabcolsep{4.5pt}
      \renewcommand\arraystretch{1.05}
      \begin{tabular}{z{45}y{20}cccccc}
         \toprule[1.0pt]
           \multicolumn{2}{c}{Method} & Backbone & $\!\!\mathcal{J}$\&$\mathcal{F}_m$$_{\!\!\!}$ $\uparrow$ & $\mathcal{J}_m$$_{\!\!\!}$ $\uparrow$ & $\mathcal{J}_r$$_{\!\!\!}$ $\uparrow$  &  $\mathcal{F}_m$$_{\!\!\!}$ $\uparrow$  & $\mathcal{F}_r$$_{\!\!\!}$ $\uparrow$  \\
          \hline
         MAST\!~\cite{lai2020mast}&\ \!\!\!\!\!\!\pub{CVPR20} & \texttt{ResNet-18}  & 54.3 & 50.7 & 58.9 & 57.8 & 64.5 \\
         CRW\!~\cite{jabri2020space}&\ \!\!\!\!\!\!\pub{NeurIPS20} & \texttt{ResNet-18}   & 55.9 & 52.3 & - & 59.6 & - \\
         DUL\!~\cite{araslanov2021dense}&\ \!\!\!\!\!\!\pub{NeurIPS21} & \texttt{ResNet-18}  &  57.0 & 53.5 & 60.4 & 60.5 & 67.6 \\
         SCR\!~\cite{son2022contrastive}&\ \!\!\!\!\!\!\pub{CVPR22} & \texttt{ResNet-18} & 59.9 & 55.9 & - & 64.0 & - \\
         LIIR\!~\cite{li2022locality}&\ \!\!\!\!\!\!\pub{CVPR22} & \texttt{ResNet-18} & 57.5 & 55.2 & 63.1 & 59.8 & 68.6\\
         \hline
         & & \texttt{ResNet-18} & \textbf{61.3} & \textbf{59.4} & \textbf{66.5} & \textbf{63.1} & \textbf{73.7}\\
         \multirow{-2}{*}{\textbf{\textsc{Ours}}} & & \texttt{ResNet-50} & \textbf{62.4} & \textbf{60.6} & \textbf{66.9} & \textbf{64.2} & \textbf{74.3} \\
         \hline
         \!\color{mygray2}RGMP\!~\cite{oh2018fast}&\ \!\!\!\!\!\!\pub{CVPR18}  & \texttt{\color{mygray2}ResNet-50} & \color{mygray2}52.9 & \color{mygray2}51.3& \color{mygray2}- & \color{mygray2}54.4 & \color{mygray2}- \\
         \!\color{mygray2}STM\!~\cite{oh2019video}&\ \!\!\!\!\!\!\pub{ICCV19} & \texttt{\color{mygray2}ResNet-50} & \color{mygray2}72.2 & \color{mygray2}69.3 & \color{mygray2}- & \color{mygray2}75.2 &\color{mygray2}-
         \\ \bottomrule[1.0pt]
      \end{tabular}
   }
\caption{\textbf{Quantitative results} (\S\ref{sec:performance})$_{\!}$~on$_{\!}$~DAVIS$_{17}$$_{\!}$~\cite{perazzi2016benchmark}$_{\!}$~\texttt{test-dev}.}
   \label{results:testdev}
   \vspace{-16pt}
\end{table}

\noindent\textbf{Performance$_{\!}$ on$_{\!}$ YouTube-VOS.$_{\!}$} We$_{\!}$ further$_{\!}$ conduct$_{\!}$ expe- riments$_{\!}$ on$_{\!}$ YouTube-VOS$_{\!}$ \texttt{val}.$_{\!}$ As$_{\!}$ shown$_{\!}$ in$_{\!}$ {Table$_{\!}$~\ref{results:ytvos}},$_{\!}$ our$_{\!}$ ap- proach,$_{\!}$ again,$_{\!}$ achieves$_{\!}$ remarkable$_{\!}$~performance,$_{\!}$ evidencing its efficacy and generalization ability across different VOS datasets.$_{\!}$ Specifically,$_{\!}$ when$_{\!}$ opting$_{\!}$ for$_{\!}$ \texttt{ResNet-18}$_{\!}$ backbone$_{\!}$ network$_{\!}$ architecture, our$_{\!}$ approach$_{\!}$ obtains$_{\!}$ \textbf{1.7}\%$_{\!}$ absolute$_{\!}$ $\mathcal{J}$\&$\mathcal{F}_m$$_{\!}$ improvement,$_{\!}$ over$_{\!}$ the$_{\!}$ current$_{\!}$ top leading method --- DUL. Moreover, with a stronger backbone --- \texttt{ResNet-50}, our approach further improves the $\mathcal{J}$\&$\mathcal{F}_m$ score to \textbf{72.4}\%, setting a new  state-of-the-art.

\noindent\textbf{Visual Comparison Results.} Fig.\!~\ref{fig:compa} depicts the visual com-

\noindent parison results of our approach and~two competitors, MAST and DUL, on two challenging videos from  DAVIS$_{17}$  \texttt{val} and YouTube-VOS \texttt{val}, respectively. We can find CRW and LIIR, as classic, correspondence-based methods, suffer from drifting errors during mask propagation; small predic- tion errors$_{\!}$ on$_{\!}$ past$_{\!}$ frames$_{\!}$ are$_{\!}$ hard$_{\!}$ to$_{\!}$ be$_{\!}$ corrected$_{\!}$ in$_{\!}$ later$_{\!}$ frames$_{\!}$ and$_{\!}$ further$_{\!}$ lead to worse results after processing more frames. This is due to their matching-based propagation strategy. In contrast, our approach generates more reasonable segments that better align object boundaries, and performs robust to small outlier predictions, hence reducing error accumulation over time. These results verify~the efficacy of our model and support our insight that encoding mask information~is crucial for self-supervised VOS. Further detailed quantitative analyses can be found in \S\ref{sec:ablative}.

\begin{table}[t]
   \resizebox{1.01\columnwidth}{!}{
      \setlength\tabcolsep{1.0pt}
      \renewcommand\arraystretch{1.05}
      \begin{tabular}{z{68}y{30}cccccc}
         \toprule[1.0pt]
           & & & & \multicolumn{2}{c}{Seen} & \multicolumn{2}{c}{Unseen} \\ \cline{5-8}
            \multicolumn{2}{c}{\multirow{-2}{*}{Method}} & \multirow{-2}{*}{Backbone} & \multirow{-2}{*}{$\mathcal{J}$\&$\mathcal{F}_m$$_{\!\!\!}$ $\uparrow$} & \ \ $\mathcal{J}_m$$_{\!\!\!}$ $\uparrow$\ \  & \ \ $\mathcal{F}_m$$_{\!\!\!}$ $\uparrow$\ \  & \ \ $\mathcal{J}_m$$_{\!\!\!}$ $\uparrow$\ \  & \ \ $\mathcal{F}_m$$_{\!\!\!}$ $\uparrow$\ \ \\ \hline
         Colorization\!~\cite{vondrick2018tracking}&\ \!\pub{ECCV18} & \texttt{ResNet-18}  & 38.9 & 43.1 & 38.6 & 36.6 & 37.4 \\
         CorrFlow\!~\cite{lai2019self}&\ \!\pub{BMVC19} & \texttt{ResNet-18}   & 46.6 & 50.6 & 46.6 & 43.8 & 45.6 \\
         MAST\!~\cite{lai2020mast}&\ \!\pub{CVPR20} & \texttt{ResNet-18}   & 64.2 & 63.9 & 64.9 & 60.3 & 67.7 \\
         CRW\!~\cite{jabri2020space}&\ \!\pub{NeurIPS20} & \texttt{ResNet-18}   & 68.7 & 67.4 & 69.1 & 65.1 & 73.2 \\
         CLTC\!~\cite{jeon2021mining}&\ \!\pub{CVPR21} & \texttt{ResNet-18}   & 67.3 & 66.2 & 67.9 & 63.2 & 71.7 \\
         DUL\!~\cite{araslanov2021dense}&\ \!\pub{NeurIPS21} & \texttt{ResNet-18}  & 69.9 & 69.6 & 71.3 & 65.0 & 73.5 \\
         LIIR\!~\cite{li2022locality}&\ \!\pub{CVPR22} & \texttt{ResNet-18}  & 69.3 & 67.9 & 69.7 & 65.7 & 73.8 \\
         \hline
         & & \texttt{ResNet-18} & \textbf{71.6} & \textbf{71.0} & \textbf{74.2} & \textbf{66.0} & \textbf{75.3}\\
         \multirow{-2}{*}{\textbf{\textsc{Ours}}} & & \texttt{ResNet-50} &  \textbf{72.4} & \textbf{71.7} & \textbf{74.6} & \textbf{67.0} & \textbf{76.2}\\
         \hline
         \!\color{mygray2}OSVOS\!~\cite{caelles2017one}&\ \!\pub{CVPR17}  & \texttt{\color{mygray2}VGG-16} & \color{mygray2}58.8 & \color{mygray2}59.8& \color{mygray2}60.5 & \color{mygray2}54.2 & \color{mygray2}60.7 \\
         \!\color{mygray2}STM\!~\cite{oh2019video}&\ \!\pub{ICCV19} & \texttt{\color{mygray2}ResNet-50} & \color{mygray2}79.4 & \color{mygray2}79.7 & \color{mygray2}84.2 & \color{mygray2}73.5&\color{mygray2} 80.9
         \\ \bottomrule[1.0pt]
      \end{tabular}
   }
\caption{\textbf{Quantitative results} (\S\ref{sec:performance}) on YouTube-VOS$_{\!}$~\cite{xu2018youtube} \texttt{val}.}
   \label{results:ytvos}
   \vspace{-16pt}
\end{table}

\subsection{Diagnostic Experiments} \label{sec:ablative}
\vspace{-1pt}
To thoroughly examine our core hypotheses and model designs, we conduct a series of ablative studies on DAVIS$_{17}$ \texttt{val}.$_{\!\!}$ The$_{\!}$ reported$_{\!}$ baselines$_{\!}$ are$_{\!}$ built$_{\!}$ upon$_{\!}$ \texttt{ResNet-18}$_{\!}$ and trained by the default setting, unless otherwise specified.

\begin{table*}[t]
\centering
   \vspace{-1.em}
   \subfloat[{loss design} \label{table:ablation1}]{
   \resizebox{0.320\textwidth}{!}{
   \setlength\tabcolsep{5pt}
   \begin{tabular}{cccc}
      \toprule[1.0pt]
      \multicolumn{1}{c}{Loss}& $\mathcal{J}$\&$\mathcal{F}_m$$_{\!\!\!}$  $\uparrow$ & $\mathcal{J}_m$$_{\!\!\!}$  $\uparrow$ & $\mathcal{F}_m$$_{\!\!\!}$  $\uparrow$  \\ \hline
       $\mathcal{L}_\text{Short}$ & 57.4 & 55.8 & 58.9 \\
       $\mathcal{L}_\text{Long}$ & 67.2 & 64.9 & 69.5 \\
       $\mathcal{L}_\text{Short}$ + $\mathcal{L}_\text{Long}$ & 68.8 & 66.7 & 70.9 \\
      $\mathcal{L}_\text{Seg}$ & 62.3 & 60.5 & 64.0\\
      \hline
      {\color{red}$\mathcal{L}_\text{Seg}$+$\mathcal{L}_\text{Short}$+$\mathcal{L}_\text{Long}$} & \textbf{74.5} & \textbf{71.6} & \textbf{77.4} \\
       \bottomrule[1.0pt]
   \end{tabular}
   }
   }
   \subfloat[{number of reference frames}\label{table:ablation2}]{%
        \resizebox{0.353\textwidth}{!}{
        \setlength\tabcolsep{8.4pt}
   \begin{tabular}{cccc}
     \toprule[1.0pt]
      \#Ref. Frame& $\mathcal{J}$\&$\mathcal{F}_m$$_{\!\!\!}$  $\uparrow$ & $\mathcal{J}_m$$_{\!\!\!}$  $\uparrow$ & $\mathcal{F}_m$$_{\!\!\!}$  $\uparrow$  \\ \hline
      First           & 68.8 & 65.7 & 71.9 \\
      First$_{\!}$~+$_{\!}$~Last 1:15 & 73.2 & 70.4 & 76.0 \\
      {\color{red}First$_{\!}$~+$_{\!}$~Last 1:20} & \textbf{74.5} & \textbf{71.6} & \textbf{77.4} \\
      First$_{\!}$~+$_{\!}$~Last 1:25 & 73.5 & 70.9 & 76.1\\
      First$_{\!}$~+$_{\!}$~Last 1:30 & 72.8 & 70.2& 75.3\\
       \bottomrule[1.0pt]
   \end{tabular}
   }
   }
   \subfloat[{number of cluster centers}\label{table:ablation3}]{
   \resizebox{0.313\textwidth}{!}{
   \setlength\tabcolsep{8.5pt}
   \begin{tabular}{cccc}
      \toprule[1.0pt]
        {\#Centroid}&$\mathcal{J}$\&$\mathcal{F}_m$$_{\!\!\!}$  $\uparrow$ & $\mathcal{J}_m$$_{\!\!\!}$  $\uparrow$ & $\mathcal{F}_m$$_{\!\!\!}$  $\uparrow$  \\ \hline
       $M=2$ & 67.5 & 65.2 & 69.8 \\
       $M=3$ & 71.6 & 69.0 & 74.2 \\
       {\color{red}$M=5$} & \textbf{74.5} & \textbf{71.6} & \textbf{77.4} \\
       $M=8$ & 72.5 & 69.6 & 75.4 \\
       $M=10$ & 70.1 & 67.3 & 72.9 \\
      \bottomrule[1.0pt]
   \end{tabular}
   }
   }
   \vspace{-5pt}

   \subfloat[{pseudo mask update}\label{table:ablation4}]{%
        \resizebox{0.251\textwidth}{!}{
        \setlength\tabcolsep{3.1pt}
   \begin{tabular}{rccc}
     \toprule[1.0pt]
     Mask update & $\mathcal{J}$\&$\mathcal{F}_m$$_{\!\!\!}$  $\uparrow$ & $\mathcal{J}_m$$_{\!\!\!}$  $\uparrow$ & $\mathcal{F}_m$$_{\!\!\!}$  $\uparrow$  \\ \hline
      No update~~ & 71.1 & 68.3 & 73.9 \\
      Per 20 epoch & 72.8 & 69.9 & 75.7\\
      Per 15 epoch & 73.9 & 70.8 & 77.0 \\
      {\color{red}Per 10 epoch} &\textbf{74.5} & \textbf{71.6} & \textbf{77.4} \\
      Per \ \ 5 epoch & 72.5 & 69.5 & 75.5 \\
      Every epoch & 69.7 & 66.7 & 72.6 \\
      \bottomrule[1.0pt]
   \end{tabular}
   }
   }
   \subfloat[{recurrent refinement}\label{table:ablation5}]{%
        \resizebox{0.308\textwidth}{!}{
        \setlength\tabcolsep{3.7pt}
   \begin{tabular}{ccccc}
     \toprule[1.0pt]
      {Round} &{$\mathcal{J}$\&$\mathcal{F}_m$$_{\!\!\!}$  $\uparrow$} &{$\mathcal{J}_m$$_{\!\!\!}$  $\uparrow$}&{$\mathcal{F}_m$$_{\!\!\!}$  $\uparrow$} & {FPS} \\ \hline
      0 & 69.7 & 67.3 & 72.1 & 1.86 \\ \hline
      1 & 72.6 & 69.8 & 75.4 & 1.84 \color{mywarning}{({-1.1\%})}  \\
      2 & 73.9 & 71.1 & 76.7 & 1.80 \color{mywarning}{({-3.2\%})}  \\
      {\color{red}3} & \textbf{74.5} & \textbf{71.6} & \textbf{77.4} & 1.77 \color{mywarning}{({-4.8\%})} \\
      4 & 74.3 & 71.2 & \textbf{77.3}& 1.73 \color{mywarning}{({-7.0\%})}  \\
      5 & 74.0 & 71.0 & 77.0 & 1.69 \color{mywarning}{({-9.2\%})} \\
        \bottomrule[1.0pt]
   \end{tabular}
   }
   }
   \subfloat[{correspondence learning schema}\label{table:ablation6}]{%
      \resizebox{0.43\textwidth}{!}{
      \setlength\tabcolsep{2.1pt}
      \begin{tabular}{lccccc}
         \toprule[1.0pt]
         {Strategy} &Loss & $\mathcal{J}$\&$\mathcal{F}_m$$_{\!\!\!}$  $\uparrow$ & $\mathcal{J}_m$$_{\!\!\!}$  $\uparrow$ & $\mathcal{F}_m$$_{\!\!\!}$  $\uparrow$ & FPS 
         \\ \hline
         \textit{photometric}& MAST$_{\!}$~\cite{lai2020mast} & 65.5 & 63.3 & 67.6 & {1.13}\\
         \textit{reconstruction}&MAST$_{\!}$~\cite{lai2020mast}+$\mathcal{L}_\text{Seg}$ & \textbf{69.0} \color{mygreen}{(\textbf{\texttt{+}3.5})} & \textbf{66.4} & \textbf{71.6} & 1.01\\\hline
         \textit{cycle-consistency} &CRW$_{\!}$~\cite{jabri2020space} & 67.6 & 64.6 & 70.6 & {1.86}\\
         \textit{tracking}&CRW$_{\!}$~\cite{jabri2020space}+$\mathcal{L}_\text{Seg}$ & \textbf{71.8} \color{mygreen}{(\textbf{\texttt{+}4.2})}& \textbf{68.3} & \textbf{75.3} & 1.77\\\hline
         {\textit{contrastive}} & $\mathcal{L}_\text{Corr}$ (ours)  & 68.8 & 66.7 & 70.9 & {1.86}\\
         \textit{matching}& {\color{red}$\mathcal{L}_\text{Corr}$+$\mathcal{L}_\text{Seg}$} & \textbf{74.5} \color{mygreen}{(\textbf{\texttt{+}5.7})} & \textbf{71.6} & \textbf{77.4} & 1.77\\
         \bottomrule[1.0pt]
      \end{tabular}
   }
   }
   \captionsetup{font=small}
   \caption{{A set of ablative studies on DAVIS$_{17}$\!~\cite{perazzi2016benchmark} \texttt{val} (\S\ref{sec:ablative}). The adopted settings are marked in {\color{red}red}.}}
   \vspace{-15pt}
   \label{ablative:merge}
\end{table*}

\noindent\textbf{Training Objective.} Our model is jointly trained for mask-embedded segmentation $\mathcal{L}_\text{Seg}$ ($_{\!}$\textit{cf}.$_{\!}$~Eq.$_{\!}$~\ref{eq:loss}) and correspondence matching $\mathcal{L}_\text{Corr}$ ($=\mathcal{L}_{\text{Short}}\!+\!\mathcal{L}_{\text{Long}}$). Table$_{\!}$~\ref{table:ablation1} analyzes the influence of different training objectives. We can find that, using $\mathcal{L}_{\text{Short}}$ or $\mathcal{L}_{\text{Long}}$ individually only yields $\mathcal{J}$\&$\mathcal{F}_m$ scores of $57.4\%$ and $67.2\%$, respectively. Their combination uplifts the performance to $68.8\%$, confirming their complementarity. However, the baseline is still weaker in comparison with current top-leading correspondence-based methods, \eg, LIIR\!~\cite{li2022locality} with 72.1\%. Moreover, when using $\mathcal{L}_\text{Seg}$ solely, the model~only achieves ${62.3\%}$. This is because, without the regularization of the correspondence learning term, $k$-means suffers from random initialization of the representation and easily return trivial solutions, \eg, fragile or massive clusters. When considering all the training goals together, performance boosts can be clearly observed, \eg, \textbf{74.5\%} in $\mathcal{J}\&\mathcal{F}_m$. Under such a scheme,  unsupervised correspondence learning makes the features informative for meaningful clustering; then the produced high-quality pseudo masks allow the model to learn to make a better use of the object mask to guide segmentation.

\noindent\textbf{Reference$_{\!}$ Frame.$_{\!}$}  As$_{\!}$ usual$_{\!}$~\cite{oh2019video,lai2020mast,cheng2021rethinking},$_{\!}$ we$_{\!}$ leverage$_{\!}$ the$_{\!}$ first frame and several previous segmented frames as well as their corresponding masks,  to support the segmentation of the current frame. Table~\ref{table:ablation2} reports the related experiments.

\noindent\textbf{$k$-means Clustering.} Next we probe the impact of the num- ber$_{\!}$ of$_{\!}$ cluster$_{\!}$ centers,$_{\!}$ \ie,$_{\!}$ $M$,$_{\!}$ in$_{\!}$ Table$_{\!}$~\ref{table:ablation3}.$_{\!}$ The$_{\!}$ best$_{\!}$ perform- ance is obtained at $M\!=\!5$, roughly equal to the obvious ob- jects number, \ie, $3_{\!}\sim_{\!}4$ on average in each training video. 

\noindent\textbf{Pseudo$_{\!}$ Mask$_{\!}$ Update.$_{\!}$} During$_{\!}$ training,$_{\!}$ our$_{\!}$ approach$_{\!}$ alter- nates$_{\!}$ between$_{\!}$~clustering$_{\!}$ based$_{\!}$ pseudo$_{\!}$ mask$_{\!}$ generation$_{\!}$ and mask$_{\!}$ guided$_{\!}$ segmentation$_{\!}$ learning.$_{\!}$~In Table$_{\!}$~\ref{table:ablation4}, we study such$_{\!}$ training$_{\!}$ strategy.$_{\!}$ `No$_{\!}$ update'$_{\!}$ means$_{\!}$ that,$_{\!}$ after$_{\!}$ the$_{\!}$~ini- tial$_{\!}$ correspondence$_{\!}$ learning$_{\!}$ stage$_{\!}$ (first$_{\!}$ 300$_{\!}$ training$_{\!}$ epochs;$_{\!\!}$ see \S\ref{sec:Id}), we create pseudo masks and use them throughout the whole joint correspondence and segmentation learning stage (last 100 epochs). This baseline achieves 71.1\% $\mathcal{J}$\&$\mathcal{F}_m$. If we improve the frequency of pseudo mask update from once to twice every 20 epochs, the score is improved to \textbf{74.5}\%. But further more frequently re-estimating the pseudo masks leads to inferior performance. We speculate that it is because, when learning with the noisy pseudo masks, it needs more  epochs to optimize the network parameters, while updating the pseudo masks too frequently will easily suffer from the impact of sub-optimal  features.

\noindent\textbf{Recurrent Refinement.} We feed our predicted masks to~the segmentation decoder $\mathcal{D}$ for iterative refinement. Table$_{\!}$~\ref{table:ablation5} reports the related results. \textit{Round 0} means we follow Eq.$_{\!}$~\ref{eq:read} to leverage $\bm{V}_{q\!}$ for mask decoding.  In \textit{Round 1}, the model follows Eq.\!~\ref{eq:refinement} to warp and refine the coarse prediction $\bar{Y}_q$ and from \textit{Round 2} onwards, we replace $\bar{Y}_q$ with the output $\hat{Y}_q$ from the prior round. As seen, after two rounds of refinement, $\mathcal{J}$\&$\mathcal{F}_m$ score is improved from 69.7\% to \textbf{74.5}\%, with only negligible delay in inference speed (\ie, {\color{mywarning}{-4.8\%}}).

\noindent\textbf{Versatility.} As our self-supervised mask embedding learning (\textit{cf.}$_{\!}$~\S\ref{sec:framework})$_{\!}$ is$_{\!}$ a$_{\!}$ general$_{\!}$ framework,$_{\!}$ it$_{\!}$ is$_{\!}$ interesting$_{\!}$~to$_{\!}$~test its efficacy with~different$_{\!}$  correspondence$_{\!}$  learning  regimes$_{\!}$  (\textit{cf.}$_{\!\!}$~\S\ref{sec:relatedwork}).$_{\!}$ In Table$_{\!}$~\ref{table:ablation6},$_{\!}$ we$_{\!}$ apply$_{\!}$~our
 mask$_{\!}$ embedding$_{\!}$ learning$_{\!}$ method$_{\!}$ to$_{\!}$ MAST$_{\!}$~\cite{lai2020mast}$_{\!}$ (reconstruction$_{\!}$ based),$_{\!}$ CRW\!~\cite{jabri2020space}$_{\!}$ (cycle-consistency$_{\!}$ based),$_{\!}$ and$_{\!}$ our$_{\!}$ correspondence$_{\!}$ learning  strategy$_{\!}$  $\mathcal{L}_\text{Corr}$$_{\!}$ (\textit{cf.}$_{\!}$~\S\ref{sec:correspondence};$_{\!}$ contrastive$_{\!}$ matching$_{\!}$ based).$_{\!}$ Impre- ssively,$_{\!}$ notable$_{\!}$ performance gains are achieved over different baselines, \eg, $\textbf{3.5\%}$ on MAST, $\textbf{4.2\%}$ on CRW, and $\textbf{5.7\%}$ on our $\mathcal{L}_\text{Corr}$, in terms of $\mathcal{J}$\&$\mathcal{F}_m$. The last column of Table$_{\!}$~\ref{table:ablation6} gives inference speed, showing the additional computational budget brought by mask embedding is negligible.

\section{Conclusions}
\vspace{-1pt}
Current$_{\!}$ solutions$_{\!}$ for$_{\!}$ self-supervised$_{\!}$ VOS$_{\!}$ are$_{\!}$ commonly built$_{\!}$ upon$_{\!}$ unsupervised$_{\!}$ correspondence$_{\!}$ matching,$_{\!}$ detached from$_{\!}$ the$_{\!}$ mask-guided,$_{\!}$ sequential$_{\!}$ segmentation$_{\!}$ nature$_{\!}$ of the$_{\!}$ task.$_{\!}$ In$_{\!}$ contrast,$_{\!}$ we$_{\!}$ devised$_{\!}$ a$_{\!}$ new$_{\!}$ framework$_{\!}$ that$_{\!}$ investi- gates$_{\!}$  both$_{\!}$  mask$_{\!}$  embedding$_{\!}$  and$_{\!}$  correspondence$_{\!}$  learning for mask propagation, in an annotation-free manner. Through space-time clustering, coherent video partitions are automa- tically generated for teaching the model to directly learn mask embedding and tracking.$_{\!}$ Meanwhile, self-supervised$_{\!}$ correspondence  learning$_{\!}$ is$_{\!}$ naturally$_{\!}$ incorporated$_{\!}$ as$_{\!}$ extra$_{\!}$ regularization. In this way, our approach successfully bri- dges the gap between fully- and self-supervised VOS mo- dels in both  performance and network architecture design.\\
{\noindent\small{\textbf{Acknowledgements.}  This work  was supported by Beijing$_{\!}$ Natural$_{\!}$ Science$_{\!}$ Foundation$_{\!}$ under$_{\!}$ Grant$_{\!}$ L191004  and the Fundamental$_{\!}$ Re- search Funds for the Central Universities (No. 226-2022-00051).}}

\clearpage

\appendix


\section{Pseudo Code} \label{sec:code}
$_{\!}$The$_{\!}$ inference$_{\!}$ mode$_{\!}$ of$_{\!}$ our$_{\!}$ method$_{\!}$ is$_{\!}$ summarized$_{\!}$ in$_{\!}$ Alg.$_{\!\!}$~\ref{alg:inference}.$_{\!}$ Note that the recurrent refinement procedure is included.

\vspace{-4pt}
\section{Analysis of Pseudo Mask Updating} \label{sec:clustering}
\begin{figure*}[t]
 \renewcommand\thefigure{S1}
\centering
      \includegraphics[width=0.99\linewidth]{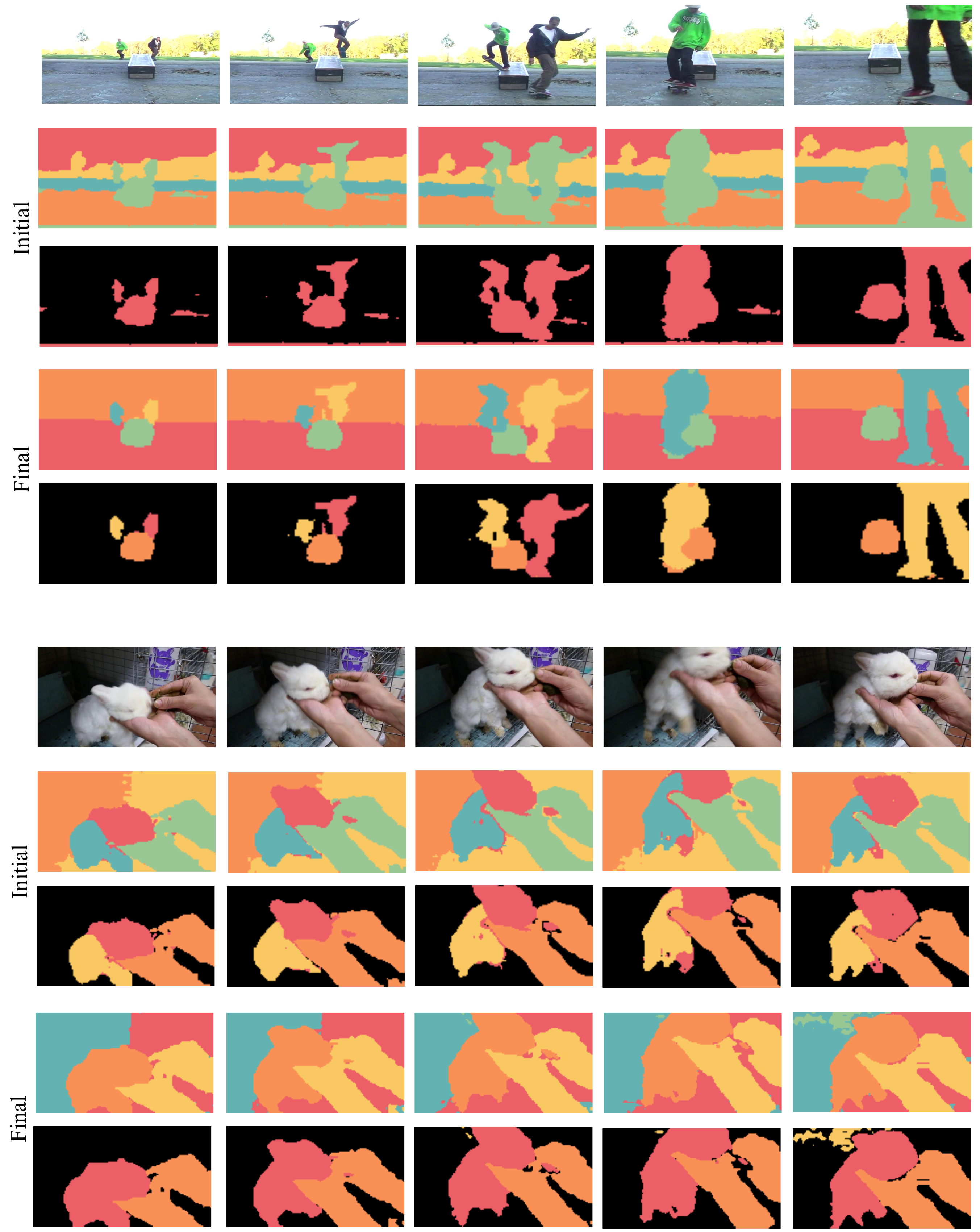}
\vspace{-3pt}
\caption{\textbf{Qualitative analysis of pseudo mask generation and update} on YouTube-VOS~\cite{xu2018youtube} \texttt{train}. `Initial': masks created right after correspondence learning; `Final': masks obtained after nine online updates (once per 10 epochs from epoch 300 to 400). The first row shows the clustering results and the second row gives the pseudo masks derived from the clustering results.}
\label{fig:pseudo1}
\vspace{-8pt}
\end{figure*}

\begin{figure*}[t]
 \renewcommand\thefigure{S2}
\centering
      \includegraphics[width=0.99\linewidth]{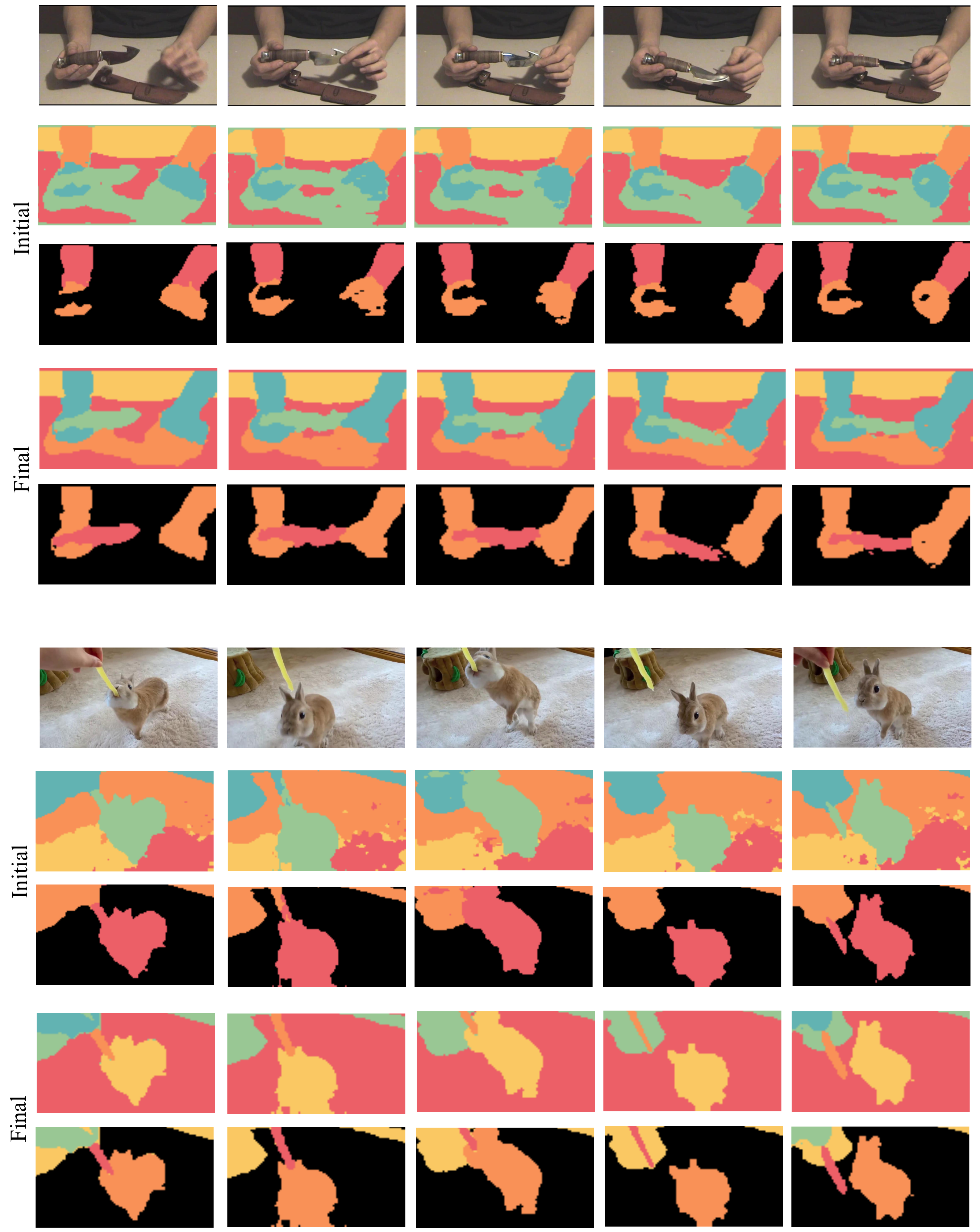}
\vspace{-3pt}
\caption{\textbf{Qualitative analysis of pseudo mask generation and update} on YouTube-VOS~\cite{xu2018youtube} \texttt{train}. `Initial': masks created right after correspondence learning; `Final': masks obtained after nine online updates (once per 10 epochs from epoch 300 to 400). The first row shows the clustering results and the second row gives the pseudo masks derived from the clustering results.}
\label{fig:pseudo2}
\vspace{-8pt}
\end{figure*}

\begin{figure*}[t]
 \renewcommand\thefigure{S3}
\centering
      \includegraphics[width=0.99\linewidth]{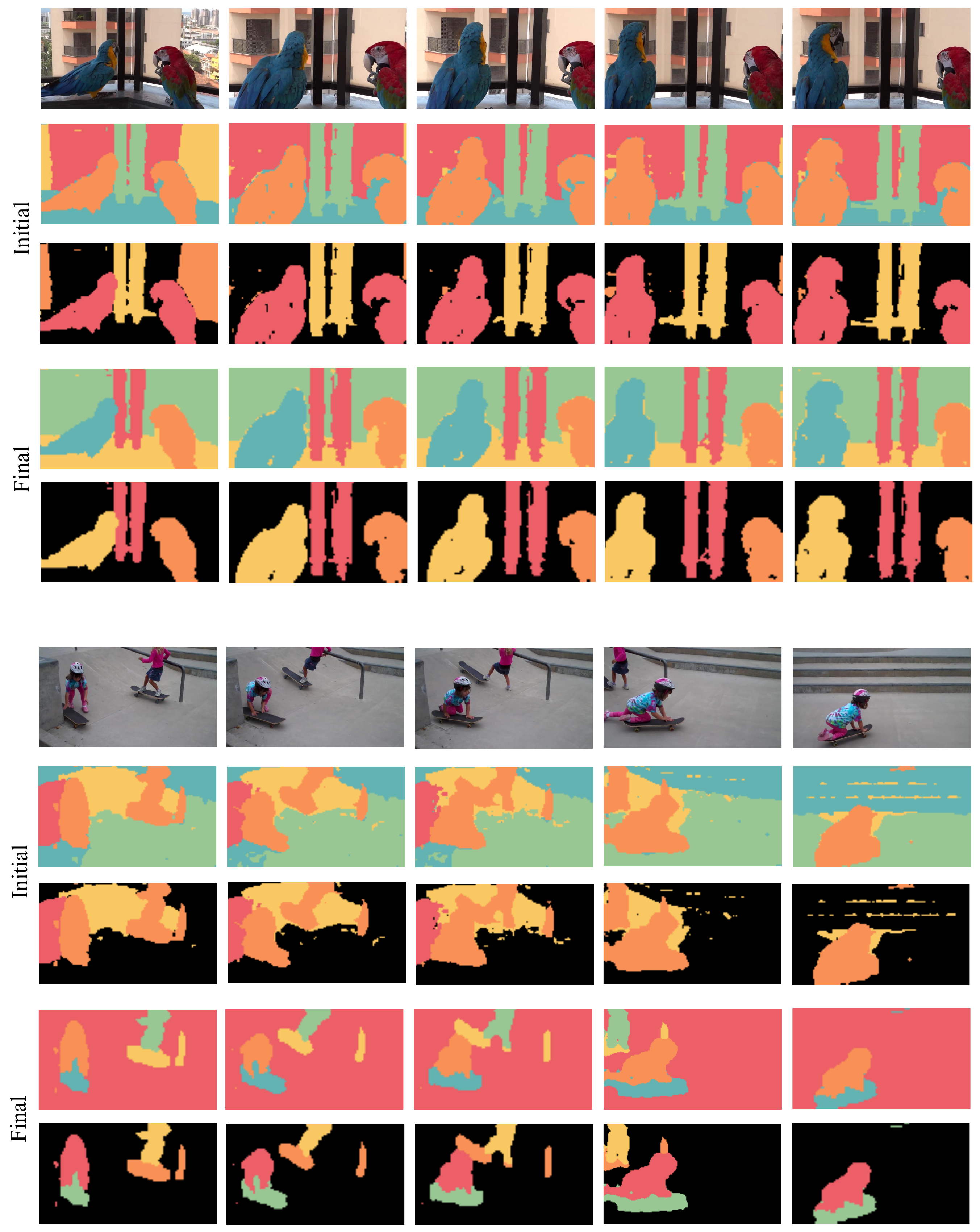}
\vspace{-3pt}
\caption{\textbf{Qualitative analysis of pseudo mask generation and update} on YouTube-VOS~\cite{xu2018youtube} \texttt{train}. `Initial': masks created right after correspondence learning; `Final': masks obtained after nine online updates (once per 10 epochs from epoch 300 to 400). The first row shows the clustering results and the second row gives the pseudo masks derived from the clustering results.}
\label{fig:pseudo3}
\vspace{-8pt}
\end{figure*}

\begin{figure*}[t]
 \renewcommand\thefigure{S4}
\centering
      \includegraphics[width=0.99\linewidth]{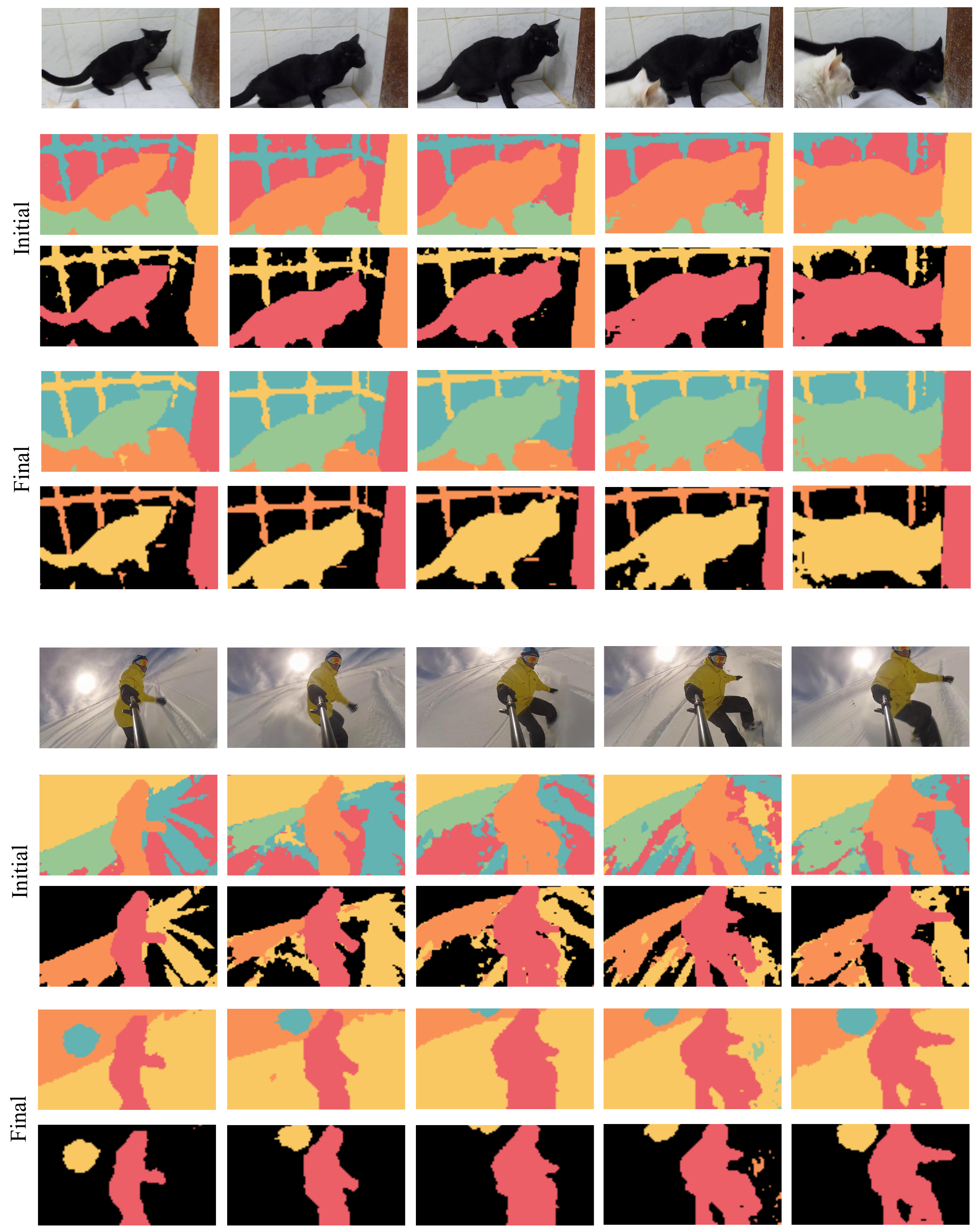}
\vspace{-3pt}
\caption{\textbf{Qualitative analysis of pseudo mask generation and update} on YouTube-VOS~\cite{xu2018youtube} \texttt{train}. `Initial': masks created right after correspondence learning; `Final': masks obtained after nine online updates (once per 10 epochs from epoch 300 to 400). The first row shows the clustering results and the second row gives the pseudo masks derived from the clustering results.}
\label{fig:pseudo4}
\vspace{-8pt}
\end{figure*}

During training, our method conducts online space-time clustering  to  progressively refine pseudo masks with  gradually improved visual representations. Fig.~\ref{fig:pseudo1}-\ref{fig:pseudo4} provide qualitative analysis of this strategy on YouTube-VOS~\cite{xu2018youtube} \texttt{train}. Here, `Initial' corresponds to the pseudo masks created right after correspondence learning, while `Final' refers to the masks that are obtained after nine online updates (once per 10 epochs from epoch 300 to 400). The first row shows the clustering results and the second row gives the pseudo masks derived from the clustering results. We can see that 1) our correspondence learning can indeed provide meaningful features for reliable clustering, leading to  satisfactory initial pseudo labels, and 2) the pseudo masks are continuously improved with online updating, \eg, back-

\vspace*{-1pt}
\begin{algorithm}
\caption{Pseudo-code for the inference mode of our approach in a PyTorch-like style}
\label{alg:inference}
\definecolor{codeblue}{rgb}{0.25,0.5,0.5}
\algcomment{\fontsize{7.2pt}{0em}\selectfont\texttt{mm}: matrix multiplication; ~~~~\texttt{normalize}: $\ell_2$ normalization; \\\texttt{cat}: concatenation; ~~~~~~~~~~~~\texttt{softmax}: row-wise softmax. }
\lstset{
  backgroundcolor=\color{white},
  basicstyle=\fontsize{7.2pt}{7.2pt}\ttfamily\selectfont,
  columns=fullflexible,
  breaklines=true,
  captionpos=b,
  escapeinside={(:}{:)},
  commentstyle=\fontsize{7.2pt}{7.2pt}\color{codeblue},
  keywordstyle=\fontsize{7.2pt}{7.2pt},
}
\begin{lstlisting}[language=python]
# I_q: query frame
# I_r: reference frames
# Y_r: reference masks of I_r
# R: number of round for recurrent refinement
# N: number of reference frames

(:\color{mycodered}{\textbf{def}}:) (:\color{mypurple}{\textbf{visual\rule{0.15cm}{0.2mm}encoder}}:)(I):
    res4, res3, res2 = (:\color{mycodegreen}{\textbf{BACKBONE}}:)(I)
    key = MLP(res4)
    key = (:\color{mypurple}{\textbf{normalize}}:)(key)
    (:\color{mycodered}{\textbf{return}}:) key, res4, res3, res2

(:\color{mycodered}{\textbf{def}}:) (:\color{mypurple}{\textbf{mask\rule{0.15cm}{0.2mm}encoder}}:)(I, Y):
    res4, _, _ = (:\color{mycodegreen}{\textbf{BACKBONE}}:)([I, Y])
    value = MLP(res4)
    (:\color{mycodered}{\textbf{return}}:) value

(:\color{mycodered}{\textbf{def}}:) (:\color{mypurple}{\textbf{inference}}:)(I_q, I_r, Y_r, R=2):
    # NHW x D'
    V_r = (:\color{mypurple}{\textbf{mask\rule{0.15cm}{0.2mm}encoder}}:)(I_r, Y_r)
    # NHW x D
    K_r, _, _, _ = (:\color{mypurple}{\textbf{visual\rule{0.15cm}{0.2mm}encoder}}:)(I_r)
    # HW x D
    K_q, res4, res3, res2 = (:\color{mypurple}{\textbf{visual\rule{0.15cm}{0.2mm}encoder}}:)(I_q)

    #===== compute the affinity (Eq.6) ======#
    # NHW x HW
    A = (:\color{mypurple}{\textbf{mm}}:)(K_r, K_q.(:\color{mypurple}{\textbf{transpose}}:)())
    A = (:\color{mypurple}{\textbf{softmax}}:)(A)

    #=== assemble support features (Eq.7) ===#
    # HW x D'
    V_q = (:\color{mypurple}{\textbf{mm}}:)(A.(:\color{mypurple}{\textbf{transpose}}:)(), V_r)

    #==== compute the coarse mask (Eq.8) ====#
    # HW x 1
    Y_q = (:\color{mypurple}{\textbf{mm}}:)(A.(:\color{mypurple}{\textbf{transpose}}:)(), Y_r)

    #======== recurrent refinement ==========#
    (:\color{mycodered}{\textbf{for}}:) _ (:\color{mycodered}{\textbf{in}}:) range(R):
    #===== predict segmentation (Eq.9) ======#
      V_q_overline = (:\color{mypurple}{\textbf{mask\rule{0.15cm}{0.2mm}encoder}}:)(I_q, Y_q)
      V_q_new = (:\color{mypurple}{\textbf{cat}}:)([V_q, V_q_overline], dim=0)
      Y_q = (:\color{mycodegreen}{\textbf{DECODER}}:)(V_q_new, res3, res2)

    (:\color{mycodered}{\textbf{return}}:) Y_q
\end{lstlisting}
\end{algorithm}

\newpage
ground are suppressed and foreground are progressively highlighted and more spatiotemporally consistent.

\vspace{-4pt}
\section{Analysis of Recurrent Refinement} \label{sec:refinement}
\begin{figure*}[t]
 \renewcommand\thefigure{S5}
\centering
      \includegraphics[width=0.99\linewidth]{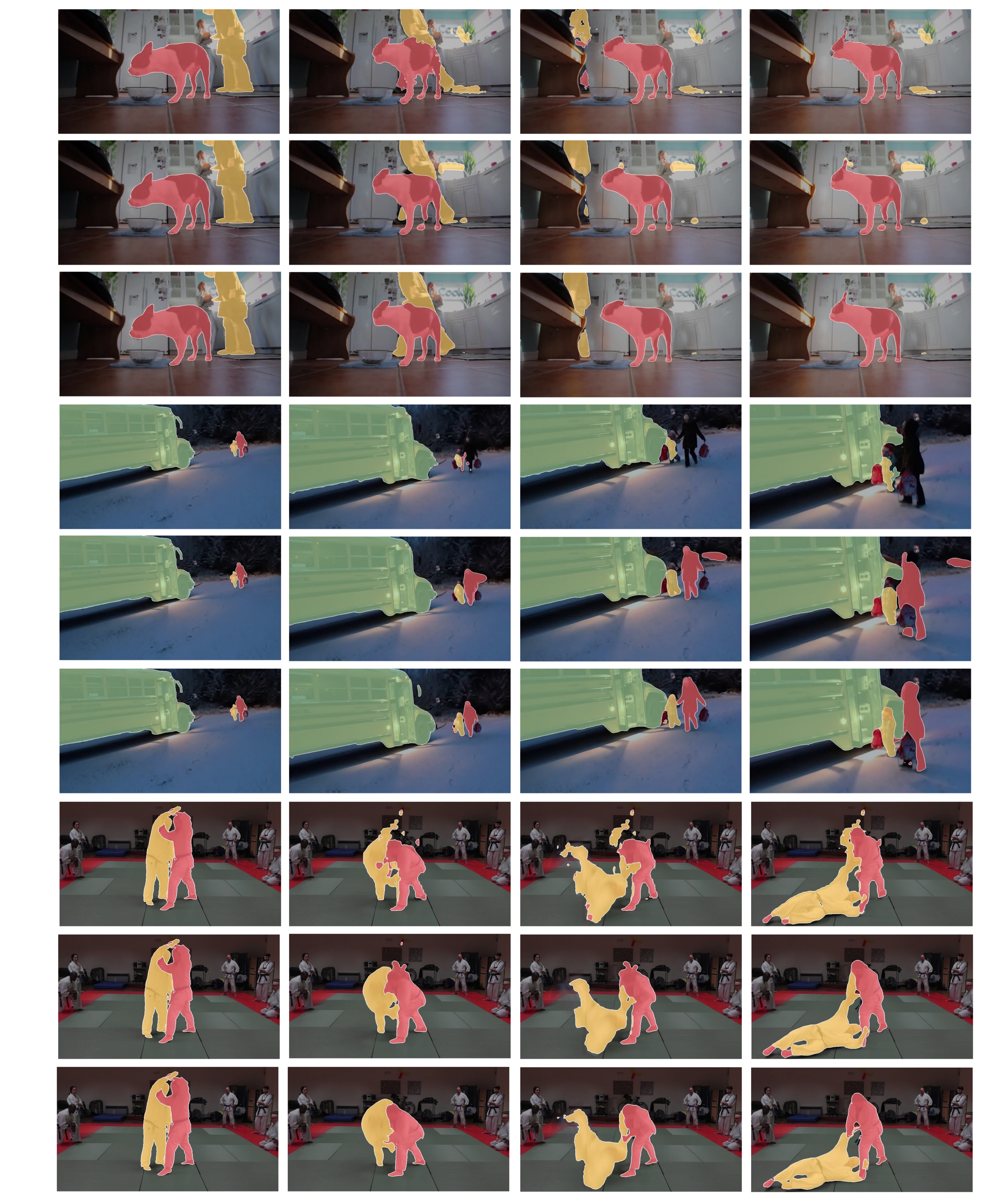}
      \put(-476.0,550){\rotatebox{90}{\textit{Round 0}}}
      \put(-476.0,486){\rotatebox{90}{\textit{Round 1}}}
      \put(-476.0,420){\rotatebox{90}{\textit{Round 2}}}
      \put(-476.0,352){\rotatebox{90}{\textit{Round 0}}}
      \put(-476.0,286){\rotatebox{90}{\textit{Round 1}}}
      \put(-476.0,222){\rotatebox{90}{\textit{Round 2}}}
      \put(-476.0,153){\rotatebox{90}{\textit{Round 0}}}
      \put(-476.0,88){\rotatebox{90}{\textit{Round 1}}}
      \put(-476.0,22){\rotatebox{90}{\textit{Round 2}}}
\vspace{-8pt}
\caption{\textbf{Qualitative analysis  of recurrent refinement} on DAVIS$_{17}$~\cite{perazzi2016benchmark} \texttt{val} and YouTube-VOS~\cite{xu2018youtube} \texttt{val}. For \textit{Round 0}, we directly leverage $\bm{V}_q$ (Eq.~{\color{red}{7}}) for mask decoding. For \textit{Round 1}, the segmentation results (\ie, $\hat{Y}_q$) are produced following Eq.~{\color{red}{9}}. For \textit{Round 2}, we replace $\overline{Y}_q$ in Eq.~{\color{red}{9}} by $\hat{Y}_q$ (in \textit{Round 1}) and subsequently conduct mask decoding to yield refined masks.}
\label{fig:round}
\vspace{-8pt}
\end{figure*}

In Fig.~\ref{fig:round}, we further analyze visual effects of recurrent refinement  over three representative sequences on DAVIS$_{17}$ \texttt{val}. For \textit{Round 0}, we directly leverage $\bm{V}_q$ (Eq.~{\color{red}{7}}) for mask decoding. For \textit{Round 1}, the segmentation results (\ie, $\hat{Y}_q$) are produced following Eq.~{\color{red}{9}}. For \textit{Round 2}, we replace $\overline{Y}_q$ in Eq.~{\color{red}{9}} by $\hat{Y}_q$ (in \textit{Round 1}) and subsequently conduct mask decoding to yield refined masks. It can be observed that the segmentation quality is progressively improved with iterative refinement, consistent with the results in Table {\color{red}{5e}}.

\vspace{-4pt}
\section{Additional Application Task} \label{sec:transfer}
We$_{\!}$ additionally$_{\!}$ test$_{\!}$ our model$_{\!}$ on$_{\!}$ the$_{\!}$ task$_{\!}$ of$_{\!}$ body$_{\!}$ part$_{\!}$ propagation.$_{\!}$ Following~\cite{wang2019learning,li2019joint,jabri2020space,xu2021rethinking,li2022locality},  we$_{\!}$ conduct experiment on$_{\!}$ VIP\!~\cite{zhou2018adaptive} benchmark dataset. It can$_{\!}$ be$_{\!}$ seen$_{\!}$ in Table~\ref{table:vip} that$_{\!}$ our$_{\!}$ method$_{\!}$ achieves$_{\!}$ the$_{\!}$ best$_{\!}$ performance.

\begin{table}
    \renewcommand\thetable{S1}
    \centering
    \resizebox{0.99\columnwidth}{!}{
        \setlength\tabcolsep{3pt}
        \renewcommand\arraystretch{1.}
        \begin{tabular}{ccccccc}
           \toprule[1.0pt]
            {Method} & TimeCycle\!~\cite{wang2019learning}  & CRW\!~\cite{jabri2020space} & CLTC\!~\cite{jeon2021mining} & VFS\!~\cite{xu2021rethinking} & LIIR\!~\cite{li2022locality} & Ours\\ \hline
            mIoU &  28.9 & 38.6 & 37.8 & 39.9 & 41.2 & \textbf{42.9} \\
           \bottomrule[1.0pt]
        \end{tabular}
    }
    \caption{Quantitative results on VIP\!~\cite{zhou2018adaptive} \texttt{test}.}
    \label{table:vip}
    \vspace{-8pt}
\end{table}

\vspace{-4pt}
\section{Additional Qualitative Results} \label{sec:vis}
\begin{figure*}[t]
 \renewcommand\thefigure{S6}
\centering
      \includegraphics[width=0.99\linewidth]{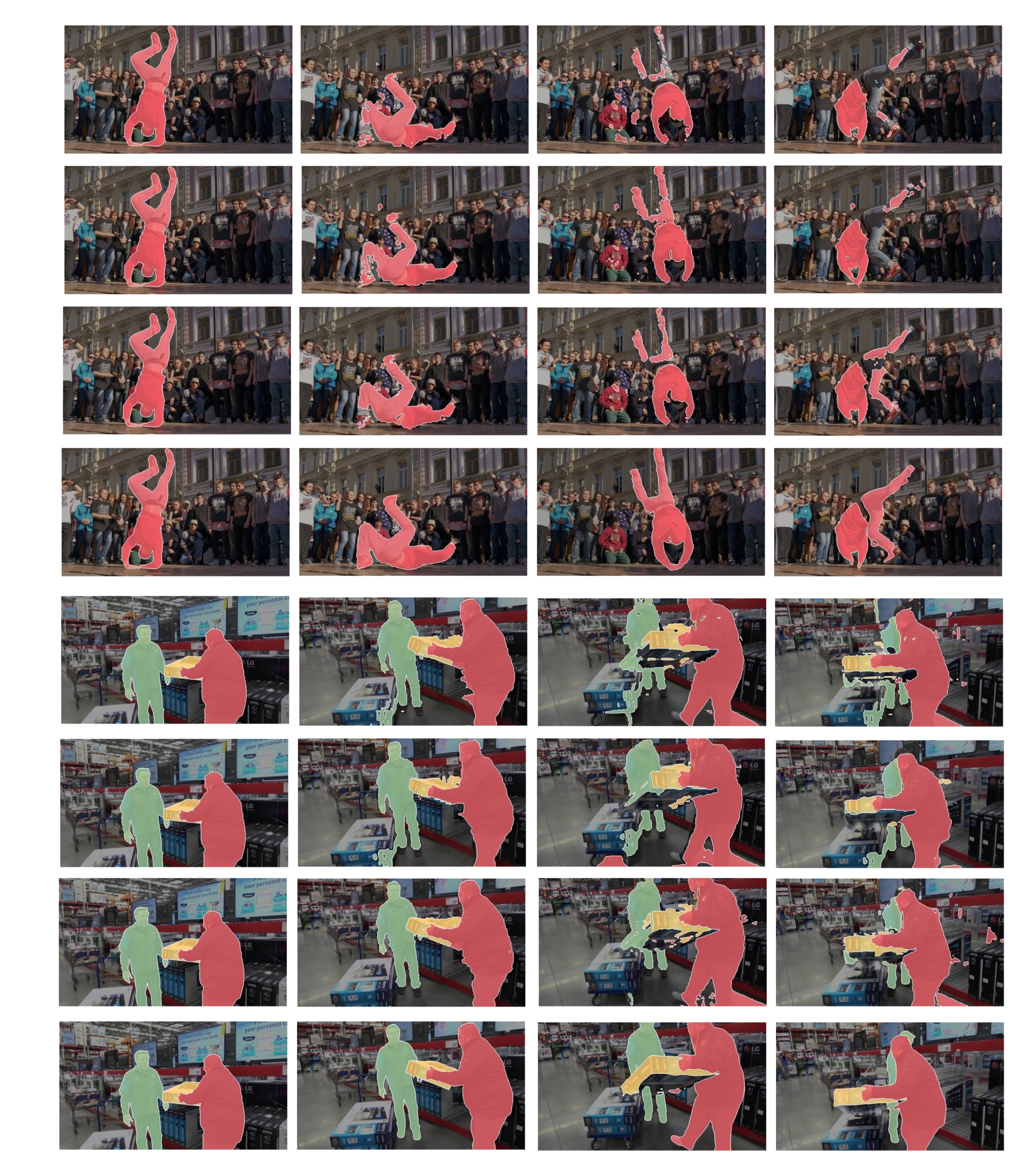}
      \put(-476.0,511){\rotatebox{90}{MAST\!~\cite{lai2020mast}}}
      \put(-476.0,445){\rotatebox{90}{CRW\!~\cite{jabri2020space}}}
      \put(-476.0,377){\rotatebox{90}{LIIR\!~\cite{li2022locality}}}
      \put(-476,315){\rotatebox{90}{\textbf{Ours}}}
      \put(-476.0,232){\rotatebox{90}{MAST\!~\cite{lai2020mast}}}
      \put(-476.0,166){\rotatebox{90}{CRW\!~\cite{jabri2020space}}}
      \put(-476.0,100){\rotatebox{90}{LIIR\!~\cite{li2022locality}}}
      \put(-476,35){\rotatebox{90}{\textbf{Ours}}}
\vspace{-3pt}
\caption{\textbf{Visual comparison results} on DAVIS$_{17}$~\cite{perazzi2016benchmark} \texttt{val}.}
\label{fig:davis1}
\vspace{-8pt}
\end{figure*}

\begin{figure*}[t]
 \renewcommand\thefigure{S7}
\centering
      \includegraphics[width=0.99\linewidth]{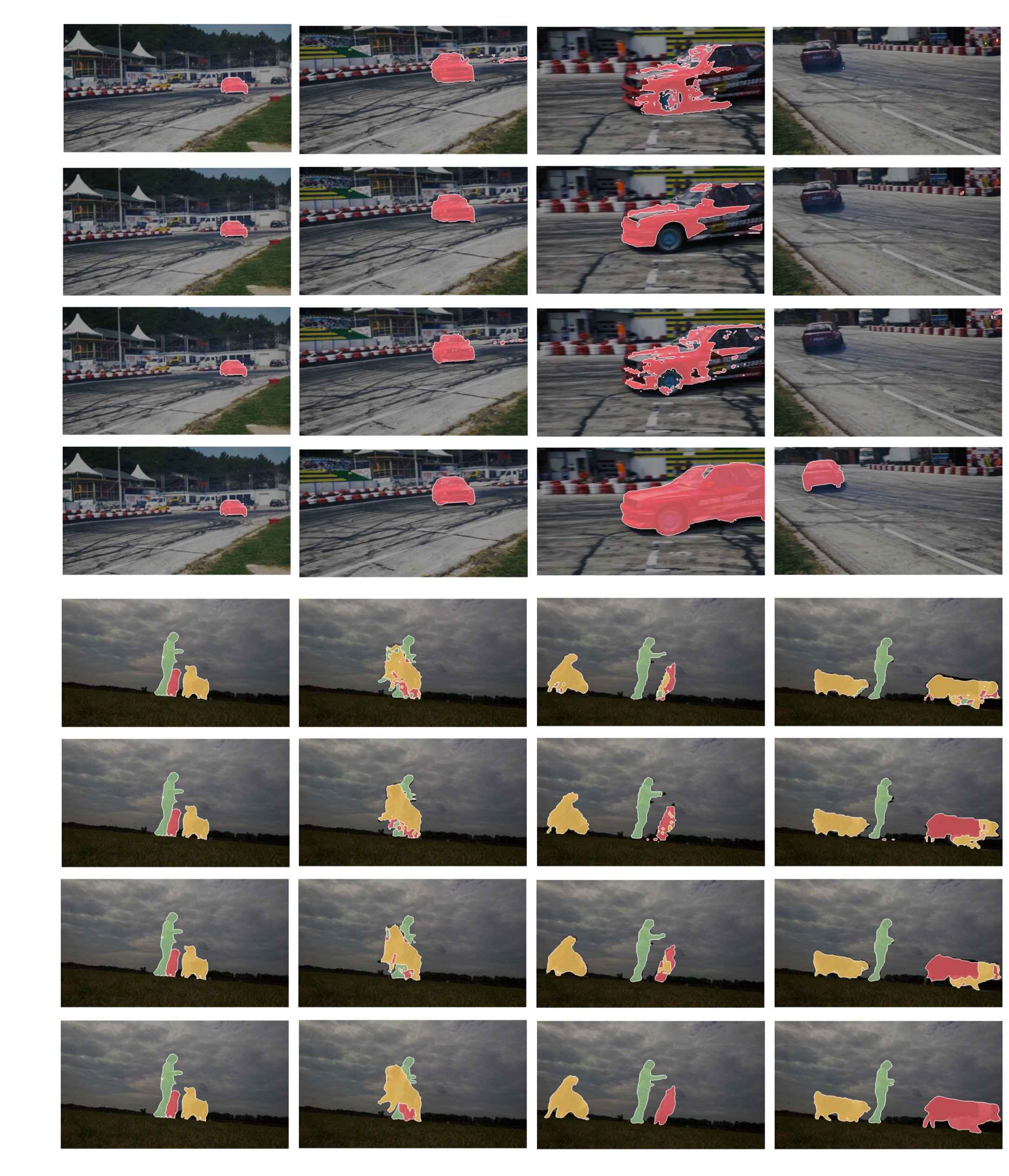}
      \put(-476.0,511){\rotatebox{90}{MAST\!~\cite{lai2020mast}}}
      \put(-476.0,445){\rotatebox{90}{CRW\!~\cite{jabri2020space}}}
      \put(-476.0,377){\rotatebox{90}{LIIR\!~\cite{li2022locality}}}
      \put(-476,315){\rotatebox{90}{\textbf{Ours}}}
      \put(-476.0,232){\rotatebox{90}{MAST\!~\cite{lai2020mast}}}
      \put(-476.0,166){\rotatebox{90}{CRW\!~\cite{jabri2020space}}}
      \put(-476.0,100){\rotatebox{90}{LIIR\!~\cite{li2022locality}}}
      \put(-476,35){\rotatebox{90}{\textbf{Ours}}}
\vspace{-3pt}
\caption{\textbf{Visual comparison results} on DAVIS$_{17}$~\cite{perazzi2016benchmark} \texttt{val}.}
\label{fig:davis2}
\vspace{-8pt}
\end{figure*}

\begin{figure*}[t]
 \renewcommand\thefigure{S8}
\centering
      \includegraphics[width=0.99\linewidth]{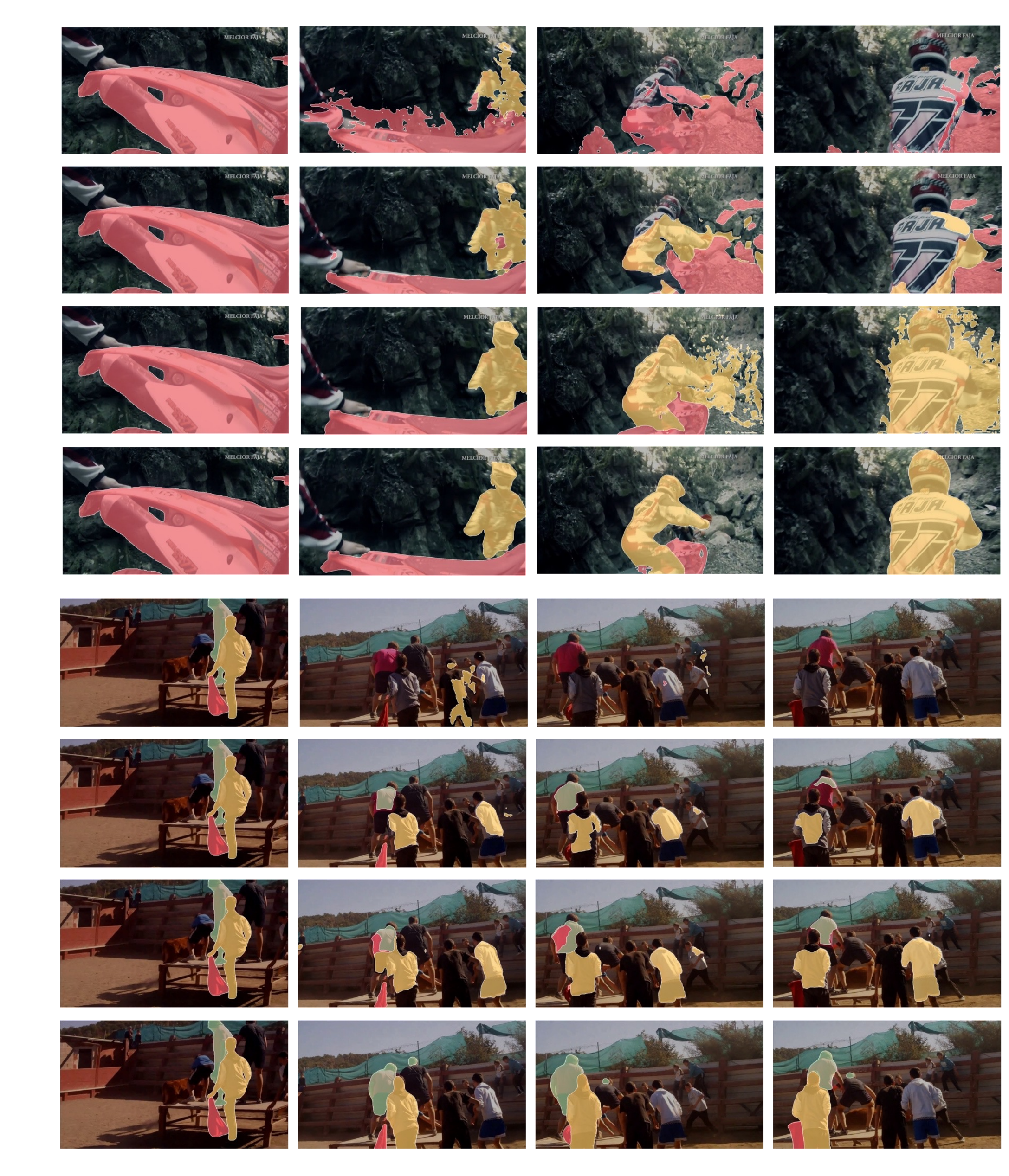}
      \put(-476.0,511){\rotatebox{90}{MAST\!~\cite{lai2020mast}}}
      \put(-476.0,445){\rotatebox{90}{CRW\!~\cite{jabri2020space}}}
      \put(-476.0,377){\rotatebox{90}{LIIR\!~\cite{li2022locality}}}
      \put(-476,315){\rotatebox{90}{\textbf{Ours}}}
      \put(-476.0,232){\rotatebox{90}{MAST\!~\cite{lai2020mast}}}
      \put(-476.0,166){\rotatebox{90}{CRW\!~\cite{jabri2020space}}}
      \put(-476.0,100){\rotatebox{90}{LIIR\!~\cite{li2022locality}}}
      \put(-476,35){\rotatebox{90}{\textbf{Ours}}}
\vspace{-3pt}
\caption{\textbf{Visual comparison results} on YouTube-VOS~\cite{xu2018youtube} \texttt{val}.}
\label{fig:ytv1}
\vspace{-8pt}
\end{figure*}

\begin{figure*}[t]
 \renewcommand\thefigure{S9}
\centering
      \includegraphics[width=0.99\linewidth]{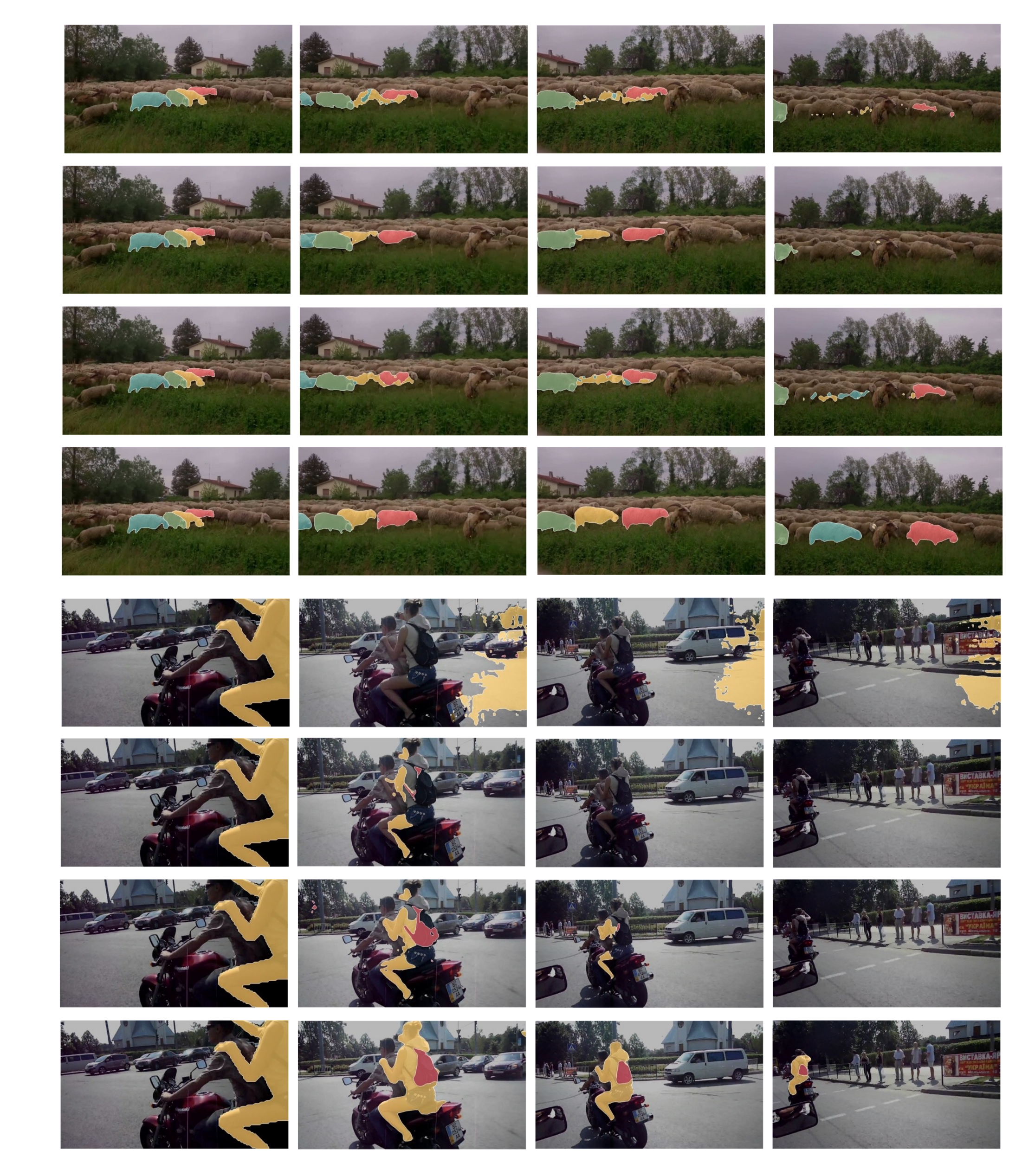}
      \put(-476.0,511){\rotatebox{90}{MAST\!~\cite{lai2020mast}}}
      \put(-476.0,445){\rotatebox{90}{CRW\!~\cite{jabri2020space}}}
      \put(-476.0,377){\rotatebox{90}{LIIR\!~\cite{li2022locality}}}
      \put(-476,315){\rotatebox{90}{\textbf{Ours}}}
      \put(-476.0,232){\rotatebox{90}{MAST\!~\cite{lai2020mast}}}
      \put(-476.0,166){\rotatebox{90}{CRW\!~\cite{jabri2020space}}}
      \put(-476.0,100){\rotatebox{90}{LIIR\!~\cite{li2022locality}}}
      \put(-476,35){\rotatebox{90}{\textbf{Ours}}}
\vspace{-3pt}
\caption{\textbf{Visual comparison results} on YouTube-VOS~\cite{xu2018youtube} \texttt{val}.}
\label{fig:ytv2}
\vspace{-8pt}
\end{figure*}

We provide more comparison results on DAVIS$_{17}$~\cite{perazzi2016benchmark} \texttt{va{}l} in Fig.~\ref{fig:davis1}-\ref{fig:davis2} and YouTube-VOS~\cite{xu2018youtube} \texttt{val} in Fig.~\ref{fig:ytv1}-\ref{fig:ytv2}, respectively. We can find that our approach suffers less from error accumulation over time, and yields consistently better results against other competitors.

\vspace{-4pt}
\section{Training Time} \label{sec:para}
The comparisons of training time are summarized in Table\!~\ref{table:time}. All experiments are conducted on one Tesla A100 GPU with \texttt{ResNet-18} backbone. $\mathcal{L}_\text{Seg}$ is involved in optimization after 300 training epochs. It can be seen that our method brings only slight training speed delay (around 8\%), while offering remarkable performance improvement.

\begin{table}
   \renewcommand\thetable{S2}
    \centering
    \resizebox{0.8\columnwidth}{!}{
        \setlength\tabcolsep{7pt}
        \renewcommand\arraystretch{1.}
        \begin{tabular}{ccc}
          \toprule[1.0pt]
            {Method} & Training time (Min/Epoch) & $\mathcal{J}$\&$\mathcal{F}_m$$_{\!\!\!}$ $\uparrow$\\ \hline
            MAST  & 26.8 & 65.5\\
            MAST + $\mathcal{L}_\text{Seg}$  & 28.7 & 69.0\\ \hline
            CRW  & 223.8 & 67.6\\
            CRW + $\mathcal{L}_\text{Seg}$ & 239.9  & 71.8\\ \hline
            $\mathcal{L}_\text{Corr}$ + (ours)  & 5.1 & 68.8\\
            $\mathcal{L}_\text{Corr}$ + $\mathcal{L}_\text{Seg}$  & 5.5 & 74.5\\
           \bottomrule[1.0pt]
       \end{tabular}
    }
    \caption{Analysis$_{\!}$ of$_{\!}$  training$_{\!}$  speed$_{\!}$  on$_{\!}$  DAVIS$_{17}$\!~\cite{perazzi2016benchmark} \texttt{val}.}
    \label{table:time}
        \vspace{-12pt}
\end{table}

\vspace{-4pt}
\section{Limitation Discussion} \label{sec:limi}
 Currently$_{\!}$  we$_{\!}$  directly$_{\!}$  leverage$_{\!}$  the$_{\!}$  $k$-means$_{\!}$  algorithm$_{\!}$  to$_{\!}$  cluster$_{\!}$  pixels.$_{\!}$  The$_{\!}$  $k$-means$_{\!}$  clustering,$_{\!}$  though$_{\!}$  simple,$_{\!}$  is$_{\!}$  less$_{\!}$  efficient$_{\!}$  compared$_{\!}$  with$_{\!}$  some$_{\!}$  more$_{\!}$  advanced$_{\!}$  ones,$_{\!}$  such$_{\!}$  as$_{\!}$  \cite{cuturi2013sinkhorn,dvurechensky2018computational} which consider clustering from the perspective of optimal transport. We leave this as a part of our future work.

\clearpage
%
%
{\small
\bibliographystyle{ieee_fullname}
\bibliography{egbib}
}

\end{document}